%% file: main.tex
\documentclass[pmlr,twocolumn,10pt,oneside,x11names,table]{jmlr} 

\usepackage{booktabs}
\usepackage{siunitx}

\usepackage[switch]{lineno}

\usepackage[nolist]{acronym}
\usepackage{xspace}  
\usepackage{inconsolata}  
\usepackage{hyperref}
\usepackage{multirow}
\usepackage{siunitx}
\usepackage[capitalise]{cleveref}
\usepackage{float}

\input{macros}

\usepackage{listings}
\lstdefinelanguage{json}{
  morestring=[b]",
  morecomment=[l]{//},
  morecomment=[s]{/*}{*/},
  morekeywords={true,false,null},
  sensitive=false,
}

\crefname{lstlisting}{Figure}{Figures}
\Crefname{lstlisting}{Figure}{Figures}

\lstdefinestyle{mypython}{
  language=Python,
  basicstyle=\ttfamily\small,
  keywordstyle=\color{blue},
  stringstyle=\color{red!60!black},
  commentstyle=\color{green!40!black},
  showstringspaces=false,
  breaklines=true,
  frame=single,
  columns=flexible,
  captionpos=b,
  abovecaptionskip=0.75em
}

\usepackage{pifont}

\theorembodyfont{\upshape}
\theoremheaderfont{\scshape}
\theorempostheader{:}
\theoremsep{\newline}

\jmlrvolume{297}
\jmlryear{2025}
\jmlrworkshop{Machine Learning for Health (ML4H) 2025} 

\title[NOVA: An Agentic Framework for Automated Histopathology Analysis and Discovery]{\textsc{NOVA}: An Agentic Framework for Automated Histopathology Analysis and Discovery}

\author{%
 \Name{Anurag J. Vaidya}\nametag{$^{1}$\thanks{Work done during internship at Microsoft.}} \Email{ajvaidya@bwh.harvard.edu}\\
 \Name{Felix Meissen$^{2}$} \Email{t-femeissen@microsoft.com}\\
 \Name{Daniel C. Castro$^{2}$} \Email{dacoelh@microsoft.com}\\
 \Name{Shruthi Bannur$^{2}$} \Email{shruthi.bannur@microsoft.com}\\
 \Name{Tristan Lazard$^{2}$} \Email{t-tlazard@microsoft.com}\\
 \Name{Drew F.\,K. Williamson$^{3}$} \Email{drew.williamson@emory.edu}\\
 \Name{Faisal Mahmood$^{1}$} \Email{faisalmahmood@bwh.harvard.edu}\\
 \Name{Javier Alvarez-Valle$^{2}$} \Email{jaalvare@microsoft.com}\\
 \Name{Stephanie L. Hyland$^{2}$} \Email{stephanie.hyland@microsoft.com}\\
 \Name{Kenza Bouzid$^{2}$}\nametag{\thanks{Corresponding author}} \Email{kenza.bouzid@microsoft.com}\\[1ex] 
 \addr {$^{1}$Mass General Brigham, Boston, USA} \\
 \addr {$^{2}$Microsoft Health Futures, Cambridge, UK} \\
 \addr {$^{3}$Emory University, Atlanta, USA}
}

\begin{document}

\begin{acronym}
    \acro{LLM}{large language model}
    \acro{AI}{artificial intelligence}
    \acro{VQA}{visual question-answering}
    \acro{VLM}{vision--language model}
    \acro{MLLM}{multimodal large language model}
    \acro{CXR}{chest X-ray}
    \acro{WSI}{whole-slide image}
    \acro{ROI}{region of interest}
    \acroplural{ROI}[ROIs]{regions-of-interest}
\end{acronym}

\maketitle
\thispagestyle{empty}
\begin{abstract}
Digitized histopathology analysis involves complex, time-intensive workflows and specialized expertise, limiting its accessibility. We introduce \ours, an agentic framework that translates scientific queries into executable analysis pipelines by iteratively generating and running Python code. \ours integrates 49 domain-specific tools (\textit{e.g.}, nuclei segmentation, whole-slide encoding) built on open-source software, and can also create new tools ad hoc. To evaluate such systems, we present \ourbenchmark, a 90-question benchmark—verified by pathologists and biomedical scientists—spanning data processing, quantitative analysis, and hypothesis testing. Unlike prior biomedical benchmarks focused on knowledge recall or diagnostic QA, \ourbenchmark demands multi-step reasoning, iterative coding, and computational problem solving. Quantitative evaluation shows \ours outperforms coding-agent baselines, and a pathologist-verified case study links morphology to prognostically relevant PAM50 subtypes, demonstrating its scalable discovery potential.
\end{abstract}

\paragraph*{Data and Code Availability}
\ourbenchmark is constructed from the TCGA Breast Invasive Carcinoma (BRCA), with \acp{WSI} and metadata obtained from the GDC portal~\citep{heath2021nci}. Additionally, data from PanopTILs~\citep{liu2024panoptic}, MoNuSeg~\citep{kumar2019multi,kumar2017dataset}, Kumar~\citep{kumar2017dataset}, and TCGA-Uniform Tumour~\citep{komura2022universal} is used. Table~\ref{tab:datasets-links} provides links to public datasets. Agent framework and benchmark code is at \url{https://github.com/microsoft/nova-agent}.

\paragraph*{Institutional Review Board (IRB)}
Proposed use of public datasets was reviewed by home institution. Under policy, use of de-identified public datasets is classified as Not Human Subjects Research [per 45§46.102(e)(1)(ii), 45§46.102(e)(5)]. Guidance and data reflection questions are provided to researchers including considerations to support representativeness, transparency and intended use.

\section{Introduction}
\label{sec:intro}

Histopathology is the gold standard for cancer diagnosis and treatment planning. The digitization of glass histology slides has allowed computational advances, such as predicting primary sites in cancers of unknown origin \citep{lu2021ai} and developing multimodal prognostic biomarkers \citep{jaume2024modeling}. However, the sheer scale and complexity of these data present a significant barrier. Effective workflows often require multi-step processing and narrow, specialized tools, creating a gap between the questions researchers want to ask and their ability to answer them without extensive programming or bioinformatics expertise.

Recent advances in \acp{LLM} and open-source histopathology tools~\citep{zhang2025standardizing} create an opportunity to bridge this gap. Instead of designing bespoke workflows for each study, an \ac{LLM}-based system equipped with domain-specific tools could autonomously generate and execute analysis pipelines in response to natural language queries. However, evaluating such flexible systems remains difficult: existing medical \ac{AI} benchmarks primarily assess text-based knowledge through multiple-choice questions~\citep{singhal2023large, nori2023capabilities} or single-image \ac{VQA} tasks on static, pre-processed images~\citep{he2020pathvqa, lau2018dataset, sun2024pathmmu}, which fail to capture the iterative reasoning, planning, and coding required for computational workflows.

To address this, we introduce \ours, an agentic framework that enables complex histopathology data analysis through a natural language interface. \ours uses a core \ac{LLM} to interpret user query, generate Python code, and orchestrate a suite of modular, custom tools for multi-step analysis directly on \acfp{WSI} and associated data. Unlike prior approaches that rely on fine-tuned models for narrow tasks such as diagnosis or \ac{VQA}~\citep{lyu2025wsi,ghezloo2025pathfinder,sun2025cpathagent}, \ours supports dynamic, interactive, and dataset-level scientific discovery without requiring instruction-fine-tuned models. It integrates 49 histopathology analysis tools (\textit{e.g.,} nuclei segmentation and classification, tissue detection, supervised classification experiments) built on trusted open software packages, making it easily extensible.

We further introduce \ourbenchmark, a benchmark of 90 questions designed to evaluate computational agents in pathology. The tasks span four categories, including  pyramidal data interrogation (\textbf{DataQA}), cellular analysis (\textbf{CellularQA}), histology \ac{ROI} understanding (\textbf{PatchQA}), and gigapixel slide-level experimentation (\textbf{SlideQA}). Each question requires multi-step reasoning, iterative coding, and computational problem solving, in addition to image captioning and knowledge recall. All questions are independently verified by both a pathologist and a biomedical scientist, ensuring clinical and scientific validity. \ourbenchmark provides a rigorous testbed for evaluating agentic systems on scientifically relevant computational tasks.

In summary, our core contributions are:
\vspace{-1mm}
\begin{enumerate}\setlength\itemsep{0em}
    \item \ours: a modular agentic framework that dynamically writes and executes Python code to build custom workflows from natural language queries, without requiring instruction-fine-tuned models;
    \item A library of 49 custom histopathology analysis tools, built on open-source software, integrated into \ours to support diverse biomedical tasks;
    \item \ourbenchmark: a computational benchmark of 90 pathologist- and scientist-verified questions for evaluating agentic workflows in pathology, released publicly for the community to extend further;
    \item Comprehensive quantitative evaluation and a pathologist-verified interactive case study using \ours to link morphological properties to prognostically relevant PAM50 molecular subtypes~\citep{parker2009supervised};
    \item Failure case analysis with examples to highlight practical issues encountered by agentic frameworks.
\end{enumerate}

\section{Related Works}
\label{sec:related-works}
Agent-based frameworks are increasingly applied to healthcare tasks. Prior work has explored multi-agent collaboration for sequential diagnosis~\citep{tu2025towards,nori2025sequential}, medical QA and VQA on benchmark datasets~\citep{kim2024mdagents,zhu2025medagentboard,he2025medorch}, and orchestration of domain-specific tools for open-ended reasoning in oncology~\citep{ferber2025development} or radiology~\citep{fallahpour2025medrax}. Other systems integrate with clinical infrastructure, such as electronic health records, to automate workflow tasks~\citep{jiang2025medagentbench}. While these approaches highlight the promise of agentic methods in medicine, they are typically text-focused and do not work directly with raw data modalities, like whole-slide images.

Several agentic systems have recently been proposed for computational pathology. \citet{lyu2025wsi} combined pretrained pathology-specific models in a pipelined ensemble for WSI classification and report generation. \citet{ghezloo2025pathfinder} and \citet{sun2025cpathagent} designed navigation-based agents that traverse WSIs with fine-tuned captioning or multimodal models for diagnosis-focused VQA. Similarly, \citet{chen2025evidence} augmented a region-level model with navigation tools for open-ended diagnosis. These approaches, however, focus narrowly on diagnostic outputs, rely on fine-tuned models, and often operate on simplified ROIs or thumbnails of WSIs rather than directly engaging with full-resolution WSIs. By contrast, \ours works natively with WSIs and associated metadata, scales beyond single slides to dataset-level tasks, and leverages modular open-source tools without requiring instruction-tuned models for orchestration.

Evaluating dynamic agentic systems requires benchmarks that move beyond static question answering to capture the complexity of multi-step reasoning and dataset-scale analysis. Most medical \ac{LLM} benchmarks evaluate text-based knowledge via multiple-choice exams or curated QA datasets~\citep{singhal2023large,nori2023capabilities,arora2025healthbench,bedi2025medhelm}. Multimodal benchmarks such as PathVQA~\citep{he2020pathvqa}, PathMMU~\citep{sun2024pathmmu}, and SlideBench-VQA~\citep{chen2025slidechat} extend to pathology but rely on static captions, automatically generated questions, or compressed slide embeddings. These often produce unanswerable or trivial questions and remain limited to single-image reasoning. More importantly, they include questions for which an image is not necessary; an LLM only baseline in~\citep{chen2025slidechat}, achieves 45\% accuracy. While \citet{chen2025evidence} evaluate an agent on whole-slide data, the focus is restricted to diagnosis and the dataset is not public. To date, no benchmark supports rigorous evaluation of computational agents performing iterative reasoning, coding, and dataset-level analysis in pathology. \ourbenchmark fills this gap by providing 90 pathologist- and scientist-verified tasks that demand multi-step workflows, tool orchestration, and hypothesis testing.

\section{\textsc{NOVA}}
\begin{figure*}[t]
    \centering
    \includegraphics[width=0.85\textwidth]{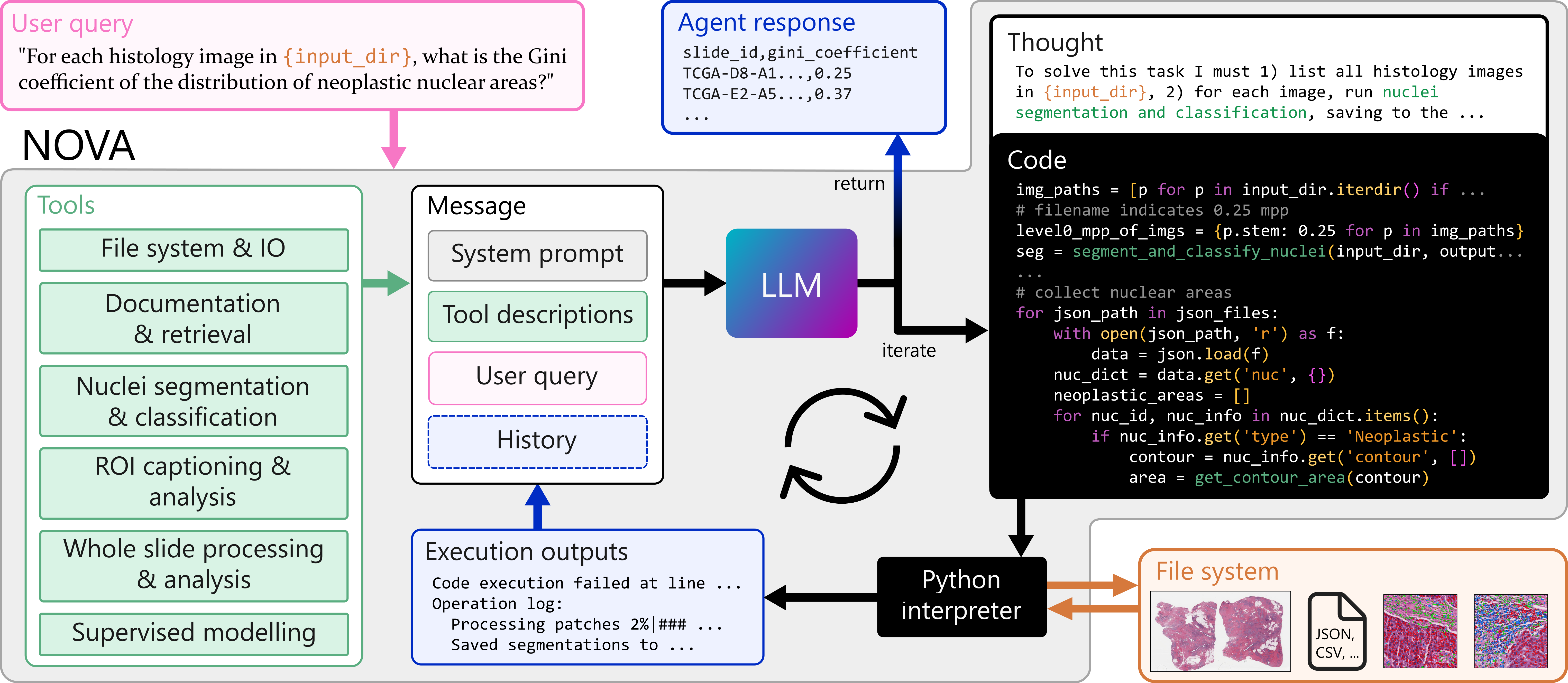}
    \caption{\ours framework. The system takes as input a user query about one or more histology images that are present on the file system. Using a collection of tools and in-built libraries, a core \ac{LLM} generates Python code to conduct multi-step data processing and analysis towards answering the user query. Code is iteratively executed and fed back into the \ac{LLM} context to enable dynamic and multi-stage action.}
    \label{fig:framework}
\end{figure*}

\ours is a modular agentic framework---based on CodeAct~\citep{wang2024executable} and developed using \texttt{smolagents}~\citep{smolagents}---that dynamically generates and executes Python code to orchestrate tool usage and answer user queries for scalable computational analysis (\Cref{fig:framework}). 

\subsection{\textsc{NOVA} Framework}
\ours is organized around three main components: (i)~a core LLM, (ii)~a Python~3.11 interpreter (\Cref{apd:python-env}) interacting with the user's file system, and (iii)~a collection of modular tools (\Cref{sec:tool-description}, \Cref{apd:tools}). User queries may include paths to data such as WSIs, ROIs, or associated metadata (\Cref{apd:user-query-format}). To process a query, \ours first dynamically constructs a system prompt with three elements: (i)~general instructions for code generation (default from \texttt{smolagents}\footnote{\href{https://github.com/huggingface/smolagents/blob/main/src/smolagents/prompts/structured_code_agent.yaml}{GitHub: \texttt{structured\_code\_agent.yaml}}}); (ii)~tool descriptions including docstrings, inputs, and outputs; and (iii)~any special instructions from the user. The combined prompt and query are passed to the core LLM, which produces structured JSON blocks containing both \texttt{thought} and \texttt{code} fields. The code is executed by the interpreter and results are fed back to the LLM for the next iteration, enabling a running memory of the reasoning process. 
The loop continues for up to 20 iterations, or until the LLM determines that the query has been fully answered. The core LLM thereby plays a vital role in \ours. These 20 iterations function as a per-task ``scratchpad", and memory is cleared after each question (except during the conversational case study).

\subsection{Tools}
\label{sec:tool-description}
The custom tools form the operational backbone of \ours. Each tool is implemented as a Python function with a clearly defined capability and is intentionally designed to be atomic rather than a multi-step workflow. This modular design ensures reusability across diverse queries and allows the \ac{LLM} to flexibly compose tools into larger, task-specific workflows. To ensure consistency and reliability, all tools follow a standardized docstring format (\Cref{fig:wsi-seg}). All 49 tools in \ours are developed using open-source computational pathology packages~\citep{zhang2025standardizing, vaidya2025molecular, zheng2025lazyslide}, making them transparent and extensible. They are organized into seven categories (\Cref{tab:histology-roi-tools,tab:dataset-check-tools,tab:dataset-pipeline-tools,tab:docs-retriever-tools,tab:nuclei-contour-tools,tab:wsi-analysis-tools,tab:wsi-classification-tools}), covering tasks from localized ROI analysis (e.g., nuclei segmentation and captioning) to whole-slide processing (e.g., tissue segmentation, patch extraction, and feature computation), as well as full supervised experiments such as training attention-based MIL models. Importantly, \ours uses LLMs that require no instruction finetuning to use the tools, lowering the barrier for adding new functionality. In addition to custom tools, \ours can access standard data science libraries (Table~\ref{tab:datascience-libs}) to autonomously generate additional tools when needed.

\section{\ourbenchmark}
\label{sec:slidequest}

\begin{figure*}
    \centering
    \includegraphics[width=0.85\textwidth]{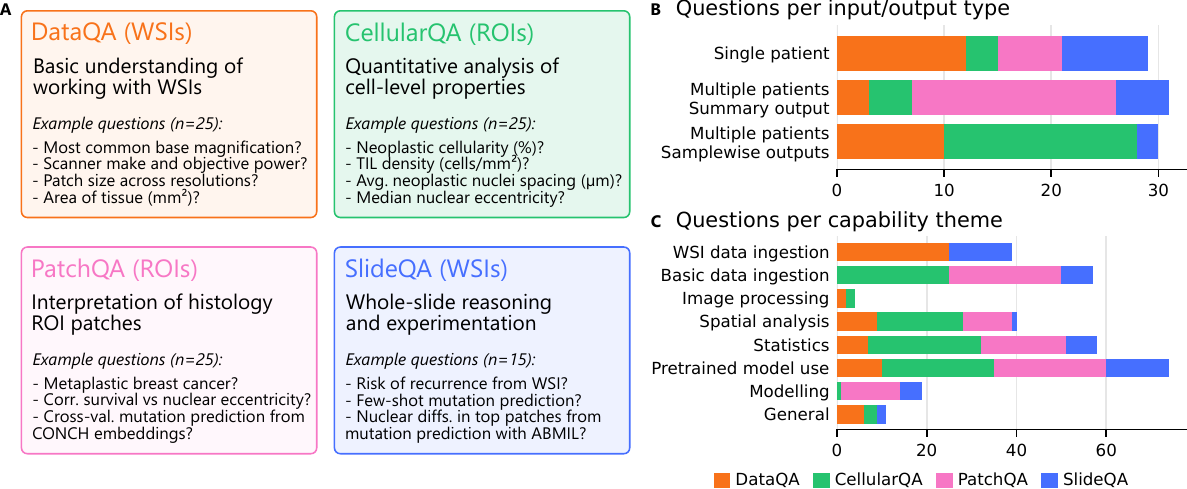}
    \caption{Overview of the \ourbenchmark benchmark.
    (A)~The four benchmark categories. Listed examples are abridged for illustration only; see full exemplars in \cref{apd:user-query-format}.
    (B)~Diversity of input and output types. Also note that DataQA and SlideQA contain \acfp{WSI}, whereas CellularQA and PatchQA operate on conventional flat images.
    (C)~Themes of capabilities required to answer the questions (full break-down in \cref{tab:capabilities}).
    }
    \label{fig:benchmark}
\end{figure*}

To evaluate the capabilities of \ours, we introduce a novel 90-question computational benchmark, \ourbenchmark. Every question is created from scratch, carefully formulated and verified to capture realistic challenges in computational pathology. \ourbenchmark is constructed entirely using publicly available data. Each question is reviewed by a computational scientist as well as a pathologist. See \cref{apd:user-query-format} for details on our unified question format and example instances. We release \ourbenchmark as an open benchmark, intended not only as a rigorous testbed for \ours but also as a template that the community can build upon and expand to other modalities. Details of datasets used to create \ourbenchmark are available in \ref{apd:detailed_datasets}.

\subsection{\ourbenchmark Categories}
\ourbenchmark is organized into categories spanning the major spatial and analytical scales of working with \acp{WSI} (\Cref{fig:benchmark}): 

\paragraph{DataQA (25 Questions):} Evaluates fundamental understanding of WSIs as a data type. Tasks include retrieving metadata from files (e.g., magnification, resolution, file format), switching between magnification levels, extracting tissue regions, and calculating basic tissue properties.

\paragraph{CellularQA (25 Questions):} Nuclei-level tasks testing the ability to segment and classify nuclei, and perform quantitative analyses such as computing cellular proportions or densities.

\paragraph{PatchQA (25 Questions):} Assesses the ability to work with histology \acfp{ROI}. Tasks include encoding ROIs using histology foundation models, classifying them, and comparing cellular properties.

\paragraph{SlideQA (15 Questions):} Evaluates WSI-level reasoning and experimentation. Spans diagnosis from gigapixel slides, training and testing supervised models, and retrieving slides based on morphological or molecular similarity.

\subsection{Task Diversity}

\ourbenchmark has a mix of questions about single or multiple patients, and with histological data as conventional flat images or pyramidal \acp{WSI}. Answers are a structured combination of binary, categorical, and numeric values, and outputs may be sample-wise or summaries over an entire dataset. We further labelled our benchmark questions according to 33 capabilities needed to answer them, grouped into 8 major themes. Regarding \emph{data ingestion}, \ourbenchmark largely demands handling pyramidal \ac{WSI} files and metadata. Quantification questions involve \emph{image processing} and precise \emph{spatial analysis} (\textit{e.g.}~physical areas/distances, morphometry). A few of the questions also need \emph{general} capabilities such as calculation and autonomous problem solving. Additionally, many tasks require \emph{using pretrained models} (for embedding, segmentation, classification) and/or \emph{training models} on the data provided or extracted embeddings. Lastly, most solutions need \emph{statistical} capabilities for e.g.~summarising outputs, cross-validating experiments, and testing hypotheses. A detailed break-down of specific capabilities is given in \cref{tab:capabilities}.

\subsection{Ground Truth and Evaluation}
\label{sec:eval}
Depending on the question, the ground-truth answers are derived from (i)~expert-provided annotations, (ii)~clinical diagnoses associated with the patients, or (iii)~human-written Python code. Multi-step questions require a combination of these approaches. Every code-derived answer is verified by a biomedical scientist to ensure correctness and reproducibility. Code to generate answers will be released with the benchmark. All baselines are instructed to produce answers as JSON files following a predefined schema in the question (\Cref{apd:user-query-format}). The answers are compared against a corresponding ground truth JSON with task-specific tolerances (details in~\Cref{apd:slidequest_eval}).

\section{Experimental Setup}

We benchmark \ours using Azure OpenAI LLM endpoints. All experiments are conducted on machines with a single NVIDIA A100 GPU within Azure ML. To account for \ac{LLM} stochasticity and to quantify variability, each experiment is repeated three times. \Cref{apd:experiment-setup} provides further details.

\subsection{Baselines}
\label{sec:baselines}

All coding baselines have access to the same Python interpreter (\cref{apd:python-env}), with an identical set of libraries (\Cref{tab:builtin-tools,tab:datascience-libs}), and are constrained to the same maximum number of iterations (20).
We compare \ours against:

\paragraph{\llmonly:} Answers queries in natural language and python code, but does not have access to a coding environment and tools.

\paragraph{\llmwpi:} Has access to a Python 3.11 environment and can execute code only once, to evaluate whether single-shot code execution is sufficient to solve tasks.

\paragraph{\llmwpir:} Can additionally refine its code over multiple steps, correcting errors along the way. Corresponds to \ours without custom tools or custom system prompt.

\section{Results}
\label{sec:overall-results}

\subsection{Performance on \ourbenchmark}
\label{sec:slidequest_perf}

\begin{figure*}[t]
\begin{center}
    \includegraphics[width=0.8\textwidth]{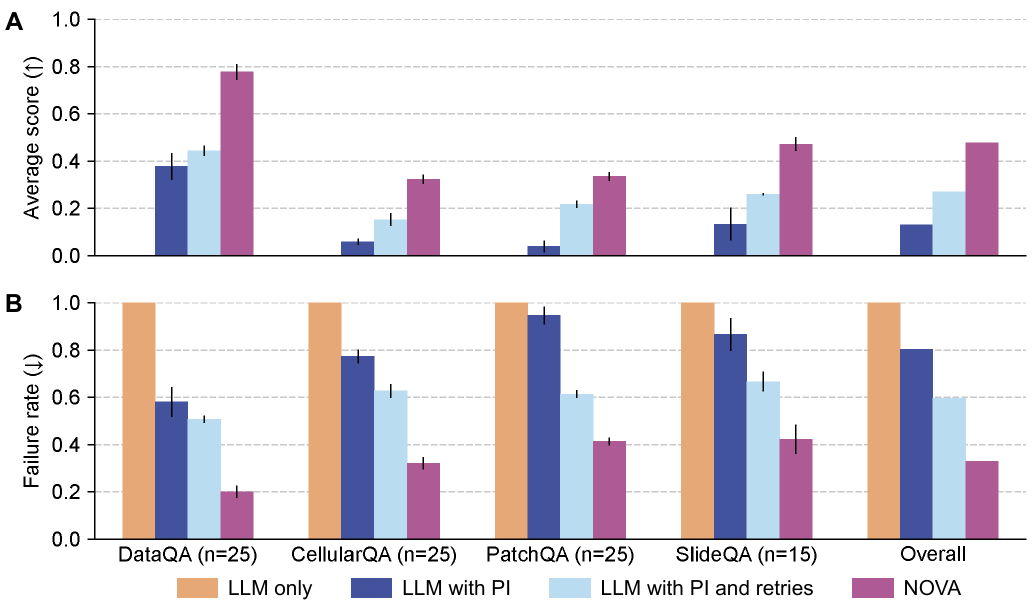}
    \caption{\textbf{A.} Average score (higher is better) on \ourbenchmark stratified by benchmark category. \textbf{B.} Failure rate (lower is better) showing the proportion of questions from \ourbenchmark on which the approach achieved a zero score. Overall is the average of each category weighted by number of questions in the category. Error bars are standard error of the mean from 3 trials. All results with \gptfourone. ``PI" stands for Python interpreter.}
    \label{fig:main_results}
    \end{center}
    
\end{figure*}

\noindent\textbf{Overall results.} \Cref{fig:main_results} demonstrates the performance of \ours against baselines across the categories of \ourbenchmark.
The \llmonly baseline achieves an average score of 0 on \ourbenchmark, confirming that the benchmark evaluates computational rather than purely linguistic ability. Adding access to a Python interpreter (\llmwpi) improves performance score to 0.154, showing that even single-shot code execution can solve a subset of tasks.
However, with \llmwpir, this score reaches an average score of 0.269 by iteratively refining code and correcting errors.
\ours achieves the highest performance on \ourbenchmark (0.477), outperforming all baselines. Averaged across all 90 benchmark questions, \ours surpasses \llmwpi by 0.323, and \llmwpir by 0.208, establishing a clear margin of improvement (\Cref{fig:main_results,tab:main-results}). \ours sequentially answered all 90 questions in 40 hours on a consumer grade GPU, which can be further sped up if running jobs in parallel asynchronously. Per category run-times are found in \Cref{apd:run-time-analysis}.

\paragraph{\textbf{Per category results.}} Stratifying \ours's performance on \ourbenchmark by category reveals substantial variation. \ours achieves its highest score on DataQA (0.777) and its lowest on CellularQA (0.323). Strong performance on DataQA is expected, as it primarily involves reading pyramidal metadata, a task well within the internal knowledge of LLMs. However, despite this understanding of WSIs, \ours exhibits a failure rate of 0.422 on SlideQA, highlighting the difficulty of computational WSI analysis. Scores below 1 were observed in CellularQA even when the correct nuclei segmentation and classification tools were used, reflecting the shortcomings of the current HoverNet~\citep{graham2019hover} model used.
We anticipate improved results as more robust tools are developed and integrated into \ours~\citep{adjadj2025towards}.

\paragraph{\textbf{Different core LLMs.}} Stronger LLMs (\gptfive vs.\ \gptfourone) improve performance on harder categories such as SlideQA (0.472 vs.\ 0.551) but show no gain or even declines on easier ones like DataQA (0.777 vs.\ 0.708). \gptfive also incurs substantially longer runtimes (e.g., averaging 47.4 vs.\ 31.2 hours for \gptfourone on SlideQA) (\Cref{fig:duration_llms}). \gptfivemini provides a strong alternative to \gptfourone, achieving the highest performance on SlideQA while matching or underperforming on the other categories (\Cref{tab:abl-llms}).

\paragraph{\textbf{Are custom tools needed?}} To understand the contribution of custom tools in our framework, we compared \ours (with custom tools) with two variants: (i) \ours (no custom tools), which measures the core LLM's ability to write tools using its internal knowledge (ii) \ours (with RAG), which does RAG on open-source computational pathology software packages to create tools based on user queries (details in~\Cref{apd:model_ablations}). \ours (with custom tools) shows clear gains over \ours (no custom tools), with use of custom tools outperforming across all categories (e.g., +0.240 on DataQA, +0.171 on CellularQA, +0.113 on PatchQA, +0.033 on SlideQA). Run time increased substantially without tools but did not improve performance: on DataQA, \ours with tools required $2.76$\,h compared to $4.20$\,h without tools, while also achieving higher performance ($0.777$ vs.\ $0.537$). Even against \ours (with RAG), the custom tool version performs better (overall performance of 0.337 with RAG and 0.477 with custom tools), indicating that RAG-based knowledge of documentation remains insufficient for effective tool creation (\Cref{tab:abl-rag}). Overall, careful manual tool design remains essential.

\subsection{Failure Case Analysis}
We manually reviewed all cases where \ours achieved a score of 0 in all runs. Failures fall into four main categories: (i)~tool limitations, (ii)~framework limitations, (iii)~ignoring existing tools or data, and (iv)~LLM fabrications. Examples of failure cases are shown in \Cref{apd:failure_analysis}.

\paragraph{\textbf{Tool limitations.}} Tool issues occurred across CellularQA, SlideQA, and PatchQA, due to incorrect output from segmentation models or image--text models, causing the final agent output to be incorrect. Moreover, tools may present results in a slightly different manner than presented in the task (for example, class names differing between the tool and task) causing the model to occasionally omit relevant classes and fabricate irrelevant ones.

\paragraph{\textbf{Framework limitations.}} The most common issue was exceeding the Python interpreter’s operation limit, a safeguard against infinite loops. This caused premature termination even when the code was correct, especially in computation-heavy tasks like CellularQA and PatchQA. The agent often retried using subsets of the data, producing incomplete or incorrect answers.

\paragraph{\textbf{Ignoring tools or data.}} In some cases, the agent recomputed values already provided by tools (e.g., convexity of tissue regions in DataQA) or rewrote code for existing functions (e.g., contour area). While not always incorrect, this behavior sometimes caused failures and often reduced efficiency. The agent also overused \texttt{try/except} blocks, skipping data that could have been recovered.

\paragraph{\textbf{Fabrications.}} Failures included fabricating data when inputs could not be loaded, or relying on simplistic heuristics (e.g., ``darker nuclei are cancerous'') instead of using tools.

\subsection{Case Study}
\begin{figure*}[!ht]
    \centering
    \includegraphics[width=0.9\textwidth]{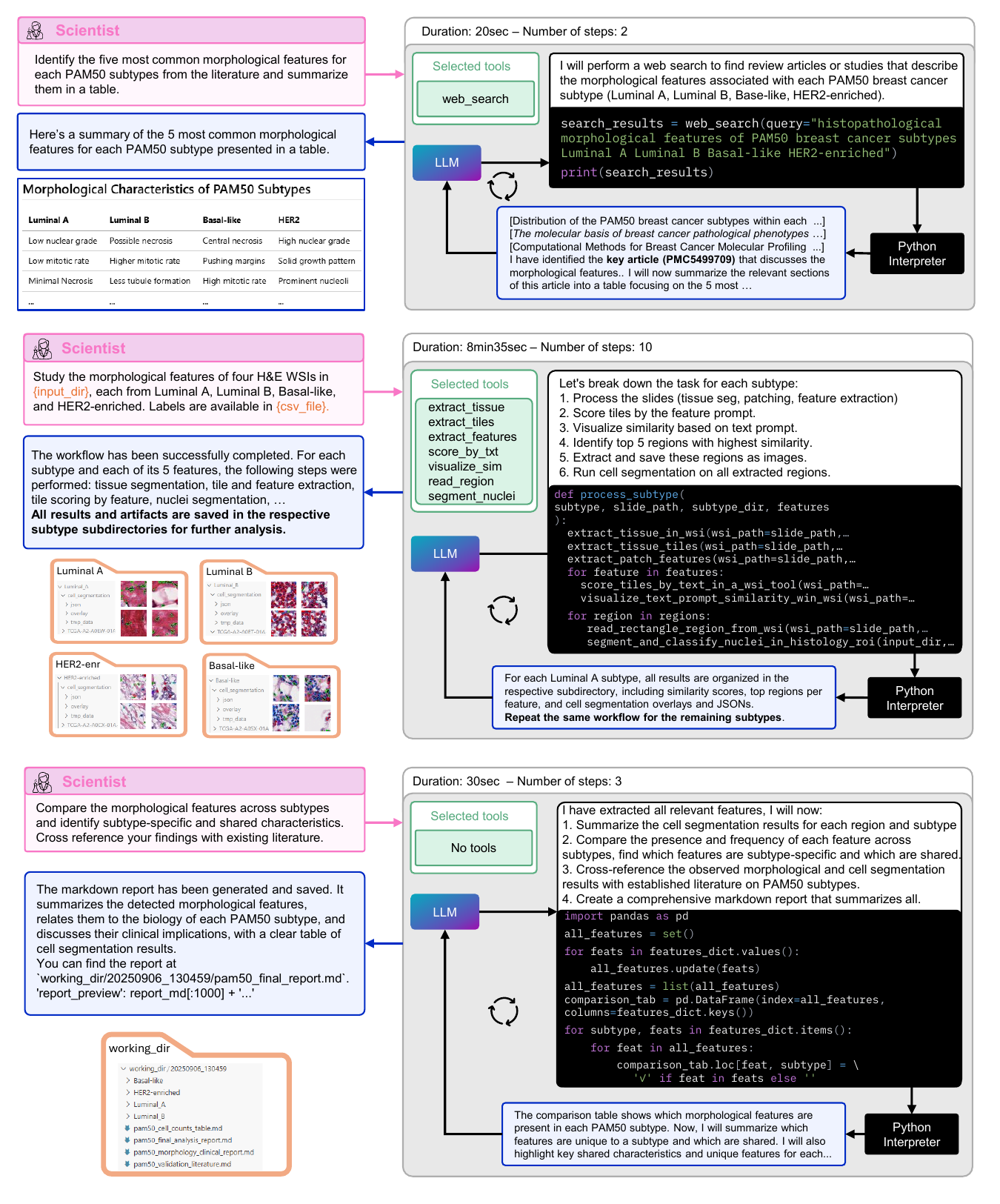}
    \caption{Case study showing the use of \ours to explore the morphological features associated with PAM50 breast cancer subtypes (Luminal A, Luminal B, Basal-like, HER2-enriched) and assess their relationship with tumour characteristics. Only the main steps are shown for illustration purposes. The final report produced by \ours is shown in \Cref{apd:final_report_part1,apd:final_report_part2}}.
    \label{fig:case_study}
\end{figure*}

We demonstrate \ours's ability to carry out a comprehensive computational pathology workflow exploring the morphological features of the four major PAM50 breast cancer subtypes: Luminal A, Luminal B, Basal-like, and HER2-enriched. \Cref{fig:case_study} shows similar to how scientists would approach this task, we first task \ours to gather relevant biological and clinical knowledge about the PAM50 subtypes from the literature. Next, we provide representative H\&E WSIs for each subtype and ask \ours to analyse them. It defines a workflow involving tissue segmentation, patch-level feature extraction, text-prompt similarity analysis to localize subtype-specific features, and nuclei segmentation. From these analyses, \ours produces a comparative report that highlights both shared and distinct characteristics of the subtypes. The results closely match known histopathological findings \citep{heng2017molecular}. For instance, Luminal A tumours show limited necrosis and abundant connective tissue, whereas Basal-like tumours exhibit extensive necrosis, inflammation, and immune infiltration. This study demonstrates \ours's ability to orchestrate complex computational workflows and derive insights by integrating multiple tools and biological knowledge, illustrating practical utility in a real-world biomedical research scenario. 

\section{Discussion}
Our findings show that strong performance on \ourbenchmark requires structured tool use and iterative coding, not natural language ability alone. Even with correct tool composition, many questions remain difficult due to current tool limitations. Modular agentic systems like \ours can readily incorporate stronger pretrained models as they emerge, and we anticipate rapid community progress on this benchmark. We also show \ours is compatible with various LLMs, where stronger models yield better results on challenging categories. Finally, we highlight that custom tools are essential; relying on LLM internal knowledge or API documentation is insufficient. Advances in automated tool creation and verification would further strengthen such systems~\citep{wolflein2025llm}.

\paragraph{\textbf{Limitations.}} The evaluation mechanism in \ourbenchmark only checks the final outputs. Hence, incorrect intermediate reasoning---including fabricated data, random guessing, or baseless tool calls---is not penalized. The framework also does not distinguish between errors arising from tool implementations and errors from the agent's use of the tools. Second, while our 49 custom tools cover diverse histopathology tasks, it is not feasible to anticipate every edge case, and the tools themselves may contain mistakes, limiting scalability. Third, reproducibility of agentic behaviour is an open challenge. The relatively high variance across runs indicates that generated pipelines may vary between executions, making it difficult to guarantee consistent outcomes. Finally, TCGA includes known limitations such as demographic biases ~\citep{vaidya2024demographic}; we hope the community adds more diverse tasks to \ourbenchmark independent on TCGA. 

\paragraph{\textbf{Conclusion.}} We introduced \ours, a coding agent framework for histopathology data analysis equipped with 49 carefully engineered tools. To rigorously evaluate its capabilities, we developed \ourbenchmark, a 90-question benchmark spanning multiple analytical scales. Across all categories, \ours outperforms coding baselines, demonstrating the necessity of domain-specific tools. Additionally, in a pathologist-verified case study linking morphology to prognostically relevant PAM50 subtypes, we demonstrate \ours's scalable discovery potential.

\paragraph{\textbf{Future Work.} }We hope to quantify the time savings with \ours for computational scientists. While we use histopathology as a proof of concept, the framework can be extended to other biomedical modalities. By releasing the framework, tools, and benchmark, we encourage the community to build on \ours by creating robust tools and \ourbenchmark by contributing new questions and tools to broaden its coverage.

\newpage
\bibliography{refs}

\onecolumn
\appendix
\nolinenumbers

\numberwithin{figure}{section}
\numberwithin{table}{section}
\numberwithin{lstlisting}{section}

\clearpage
\begin{center}
    {\LARGE \textbf{Supplementary Material}} \\[2em]
    {\large \textbf{NOVA: An Agentic Framework for Automated Histopathology Analysis and Discovery}} \\[2em]
\end{center}

We provide supplementary information \ours framework, \ourbenchmark, additional results, run time analysis:

\begin{enumerate}
    \item \textbf{Section}~\ref{apd:detailed_datasets}: Details of datasets used for \ourbenchmark.
    \item \textbf{Section}~\ref{apd:data_links}: Links to public datasets used to construct \ourbenchmark.
    \item \textbf{Section}~\ref{apd:python-env}: Python~3.11 execution environment, allowed libraries, and operation limits.
    \item \textbf{Section}~\ref{apd:tools}: Catalog of \ours\ tools (descriptions and categories).
    \item \textbf{Section}~\ref{apd:user-query-format}: Unified \ourbenchmark\ question schema with exemplars across categories.
    \item \textbf{Section}~\ref{apd:benchmark_details}: Details on \ourbenchmark question types.
    \item \textbf{Section}~\ref{apd:experiment-setup}: Experimental setup, including \ours\ configuration, LLM variants/parameters, and the experiment runner.
    \item \textbf{Section}~\ref{apd:slidequest_results}: Results of baseliens and \ours on \ourbenchmark.
    \item \textbf{Section}~\ref{apd:run-time-analysis}: Runtime vs.\ accuracy analyses across baselines and core LLMs.
    \item \textbf{Section}~\ref{apd:model_ablations}: Ablations of \ours on \ourbenchmark.
    \item \textbf{Section}~\ref{apd:failure_analysis}: Failure analysis with representative cases (tool limitations, recomputation, and other modes).
\end{enumerate}

\clearpage
\section{Datasets}
\label{apd:detailed_datasets}
Questions in \ourbenchmark are drawn from multiple publicly available datasets, all derived from TCGA WSIs but annotated at different levels of granularity. For \textbf{CellularQA}, we used three nuclei-level annotation datasets: MoNuSeg~\citep{kumar2019multi,kumar2017dataset} (51 images), Kumar~\citep{kumar2017dataset} (30 images), and PanopTILs~\citep{liu2024panoptic}. Each contains expert-verified nuclei boundaries and classifications. For PanopTILs, we restricted to the training set, which provides larger annotated regions of interest (ROIs) compared to the test set. We further limited PanopTILs to ROIs containing neoplastic and/or immune cells and only included patients with overall survival data. After filtering, we obtained 589 ROIs from PanopTILs (111 patients). For \textbf{DataQA} and \textbf{SlideQA}, we retrieved WSIs for the patients included above. When multiple WSIs existed for a patient, we randomly selected one, yielding 151 WSIs in total. For \textbf{PatchQA's} diagnostic capability, we used six BRCA patients (2 invasive ductal, lobular, and metaplastic) from TCGA Uniform tumour (UT) dataset~\citep{komura2022universal}, which provides tumour-level ROIs. We retained only three tumour ROIs per patient from the TCGA-UT dataset, randomly sampled when more than three were available. While \ourbenchmark is developed using breast cancer as an exemplar, we strongly encourage the community to extend this benchmark framework to other disease types.

\clearpage
\section{Case study final report}

\begin{figure*}[!ht]
    \centering
    \includegraphics[width=0.85\textwidth, page=1]{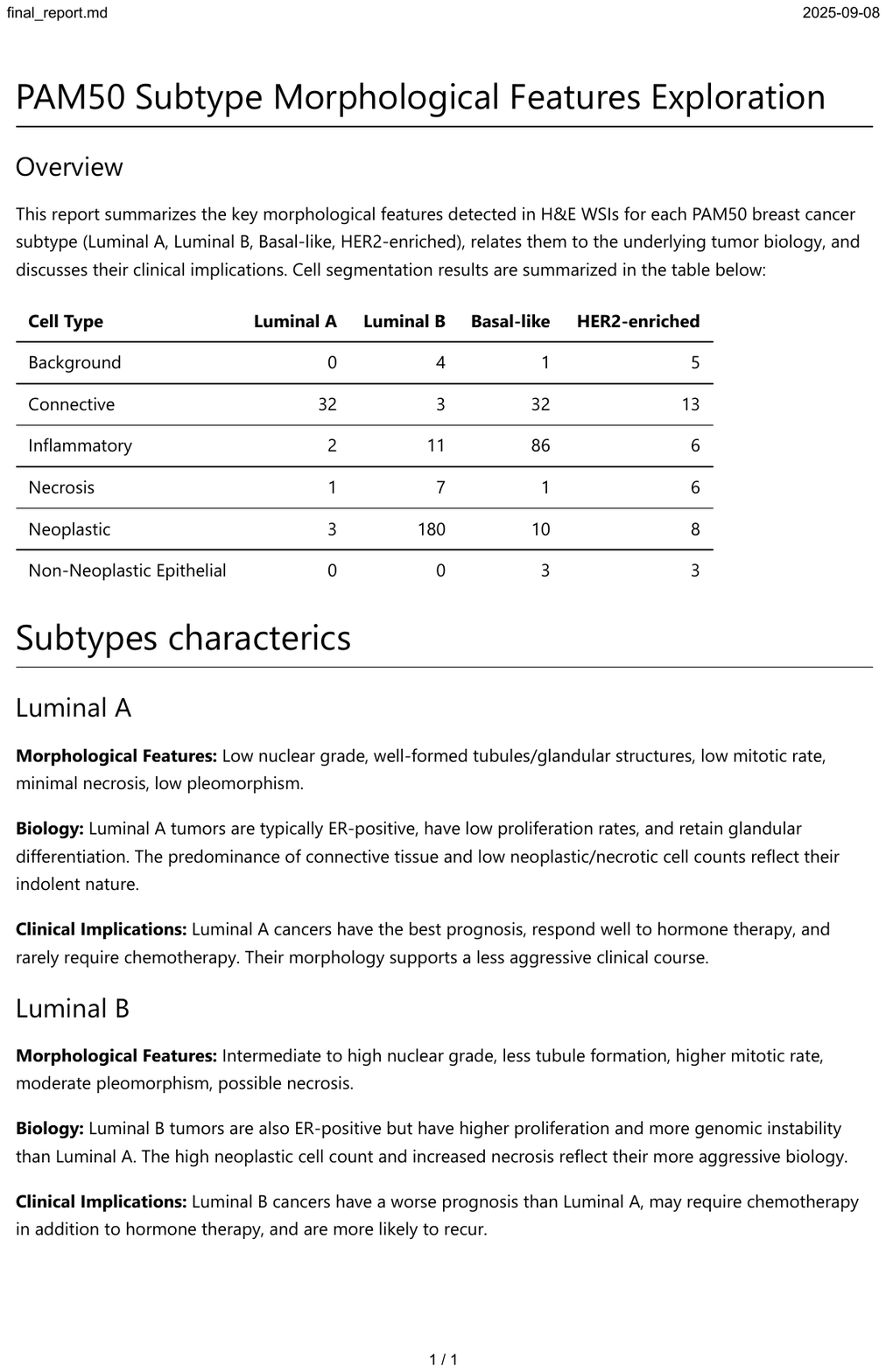}
    \caption{Final analysis markdown report (Part 1 of 2) produced by \ours for the exploration of the morphological features associated with molecular PAM50 breast cancer subtypes.}
    \label{apd:final_report_part1}
\end{figure*}

\clearpage

\begin{figure*}[!ht]
    \centering
    \includegraphics[width=0.85\textwidth, page=1]{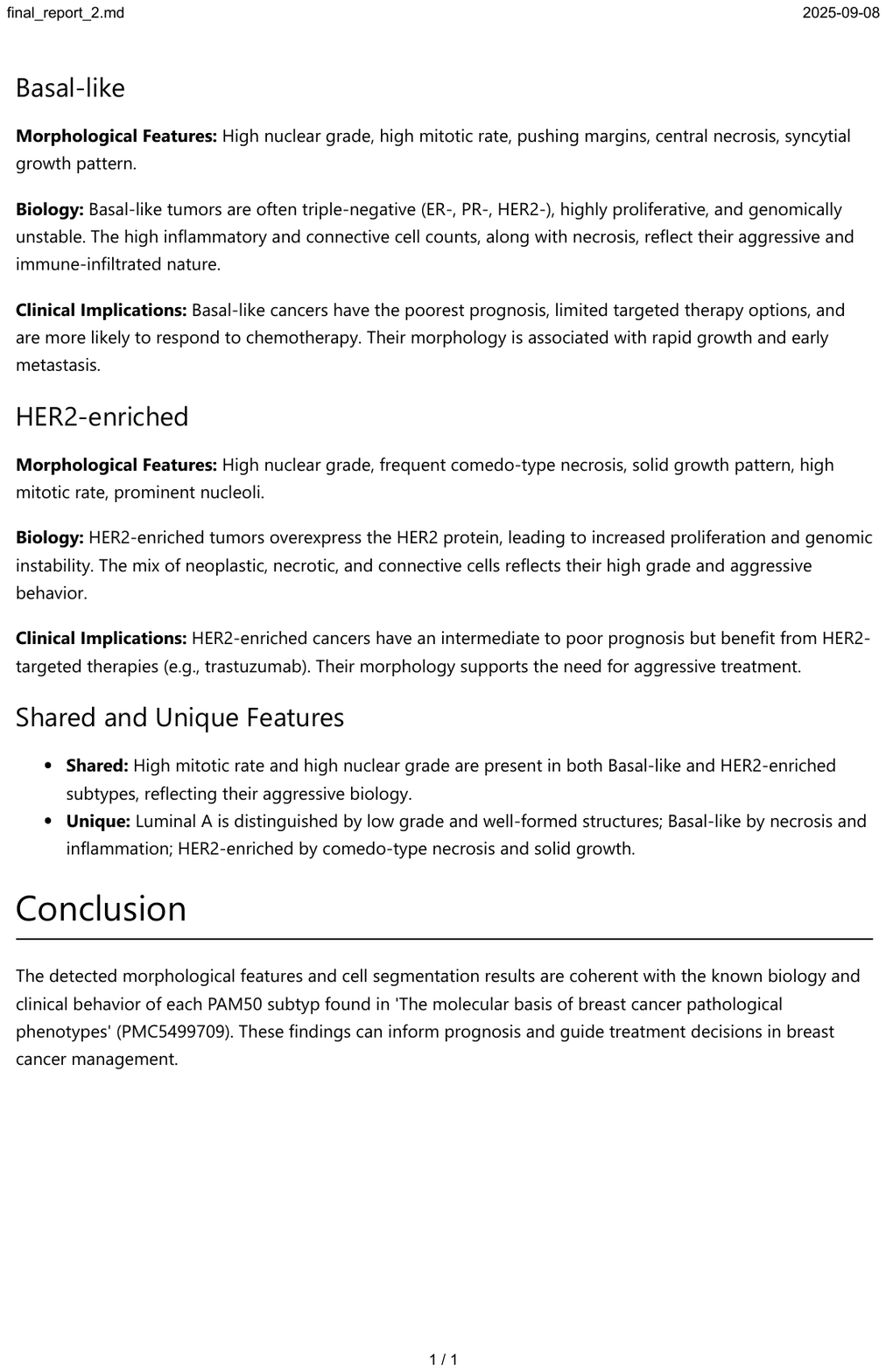}
    \caption{Final analysis markdown report (Part 2 of 2) produced by \ours for the exploration of the morphological features associated with molecular PAM50 breast cancer subtypes.}
    \label{apd:final_report_part2}
\end{figure*}

\clearpage
\section{Links to datasets}
\label{apd:data_links}
\begin{table}[h]
\centering
\caption{Public datasets used to construct \ourbenchmark.}
\label{tab:datasets-links}
\vspace{0.5em}
\begin{tabular}{ll}
\toprule
\textbf{Dataset} & \textbf{Link} \\
\midrule
TCGA BRCA (WSI + metadata) & \href{https://portal.gdc.cancer.gov/}{https://portal.gdc.cancer.gov/} \\
PanopTILs & \href{https://sites.google.com/view/panoptils/}{https://sites.google.com/view/panoptils/} \\
MoNuSeg & \href{https://monuseg.grand-challenge.org/Data/}{https://monuseg.grand-challenge.org/Data/} \\
Kumar & \href{https://drive.google.com/drive/folders/1bI3RyshWej9c4YoRW-\_q7lh7FOFDFUrJ}{Google Drive link} \\
TCGA-Uniform Tumour (TCGA-UT) & \href{https://huggingface.co/datasets/dakomura/tcga-ut}{https://huggingface.co/datasets/dakomura/tcga-ut} \\
\bottomrule
\end{tabular}
\end{table}

\section{Python environment}
\label{apd:python-env}
A local Python~3.11 environment with basic modules and data science libraries is used for code execution and file system operations. The interpreter automatically raises an error if the code generated by the LLM imports packages outside the allowed set. Security-sensitive packages such as \texttt{os}, which enable unrestricted access to the file system, are disallowed by default. A maximum number of $10^7$ code operations is allowed to avoid infinite loops. Packages and libraries available in the interpreter are listed in~\Cref{tab:builtin-tools,tab:datascience-libs}.  

\begin{table}[h]
\centering
\caption{Default Python packages available to all baselines}
\label{tab:builtin-tools}

\begin{tabular}{ll}
\toprule
\textbf{Category} & \textbf{Modules} \\
\midrule
Data structures and iteration      & collections, itertools \\
Date and time handling             & datetime, time \\
Mathematical functions             & math \\
Random number generation           & random \\
Text and regular expressions       & re \\
Statistical utilities              & stat, statistics \\
Data structures (queues)           & queue \\
Unicode utilities                  & unicodedata \\
Data serialization and file paths  & json, pathlib \\
\bottomrule
\end{tabular}
\end{table}

\begin{table}[h]
\centering
\caption{Additional Python data science libraries available to all baselines}
\label{tab:datascience-libs}

\begin{tabular}{ll}
\toprule
\textbf{Category} & \textbf{Libraries} \\
\midrule
Numerical computing                & numpy, scipy \\
Data analysis and statistics       & pandas, statsmodels \\
Machine learning and survival analysis & sklearn, sksurv, lifelines \\
Visualization                      & matplotlib, seaborn \\
Image processing and computer vision & cv2, skimage, PIL, openslide \\
Deep learning                      & torch, torchvision \\
Graphs and single-cell analysis    & networkx, scanpy \\
Data storage and serialization     & zarr, h5py, pickle \\
Geospatial and spatial data        & geopandas, spatialdata, shapely \\
Utilities (I/O, file ops, progress bars) & pathlib, shutil, tqdm \\
\bottomrule
\end{tabular}
\end{table}

\clearpage
\section{\textsc{NOVA} tools}
\label{apd:tools}

The core tools within NOVA for tasks like cell segmentation, feature extraction, metadata parsing, and region-of-interest analysis are applicable to histopathology images from any tissue.

\begin{table}[h]
\centering
\caption{Histology ROI captioning and analysis tools available in \ours.}
\label{tab:histology-roi-tools}
\vspace{0.5em}
\resizebox{\linewidth}{!}{%
\begin{tabular}{p{0.7\linewidth} p{0.48\linewidth}}
\toprule
\textbf{Tool Name} & \textbf{Description} \\
\midrule
\texttt{caption\_single\_histology\_image\_tool} & Generate a descriptive caption for a single histology image. \\
\texttt{caption\_and\_summarize\_set\_of\_histology\_images\_tool} & Caption multiple histology images and provide a summary. \\
\texttt{score\_single\_histology\_image\_using\_text\_tool} & Score a histology image based on text-based criteria. \\
\texttt{encode\_histology\_roi\_tool} & Encode histology region of interest into vector representation. \\
\bottomrule
\end{tabular}
}

\caption{Dataset processing check tools available in \ours.}
\label{tab:dataset-check-tools}
\vspace{0.5em}
\resizebox{\linewidth}{!}{%
\begin{tabular}{p{0.7\linewidth} p{0.48\linewidth}}
\toprule
\textbf{Tool Name} & \textbf{Description} \\
\midrule
\texttt{dataset\_of\_wsi\_get\_valid\_slide\_paths\_tool} & Get valid WSI file paths from a directory with optional extension filtering. \\
\texttt{dataset\_of\_wsi\_check\_tissue\_segmentation\_exists\_tool} & Check if tissue segmentation files exist for a dataset of WSIs. \\
\texttt{dataset\_of\_wsi\_check\_patch\_coordinates\_exist\_and\_schema\_tool} & Check if patch coordinate files exist and validate their schema for a dataset of WSIs. \\
\texttt{dataset\_of\_wsi\_check\_patch\_features\_exist\_and\_schema\_tool} & Check if patch feature files exist and validate their schema for a dataset of WSIs. \\
\texttt{dataset\_of\_wsi\_check\_slide\_features\_exist\_and\_schema\_tool} & Check if slide-level feature files exist and validate their schema for a dataset of WSIs. \\
\bottomrule
\end{tabular}
}

\caption{Dataset processing pipeline tools available in \ours.}
\label{tab:dataset-pipeline-tools}
\vspace{0.5em}
\resizebox{\linewidth}{!}{%
\begin{tabular}{p{0.7\linewidth} p{0.48\linewidth}}
\toprule
\textbf{Tool Name} & \textbf{Description} \\
\midrule
\texttt{dataset\_of\_wsi\_tissue\_segmentation\_tool} & Perform tissue segmentation on a dataset of WSI files. \\
\texttt{dataset\_of\_wsi\_patch\_coordinate\_extraction\_tool} & Extract patch coordinates from tissue regions in a dataset of WSIs. \\
\texttt{dataset\_of\_wsi\_patch\_features\_extraction\_tool} & Extract patch-level features from a dataset of WSIs using foundation models. \\
\texttt{dataset\_of\_wsi\_slide\_features\_extraction\_tool} & Extract slide-level features from patch features for a dataset of WSIs using slide encoders like TITAN~\citep{ding2025multimodal}, MADELEINE~\citep{jaume2024multistain}, and PRISM~\citep{shaikovski2024prism}. \\
\texttt{dataset\_of\_wsi\_create\_score\_heatmap\_tool} & Create score heatmaps overlaid on WSIs for visualization and analysis. \\
\bottomrule
\end{tabular}
}

\caption{Documentation retriever tools available in \ours.}
\label{tab:docs-retriever-tools}
\vspace{0.5em}
\resizebox{\linewidth}{!}{%
\begin{tabular}{p{0.7\linewidth} p{0.48\linewidth}}
\toprule
\textbf{Tool Name} & \textbf{Description} \\
\midrule
\texttt{trident\_docs\_retriever} & Search and retrieve information from Trident documentation for WSI processing. \\
\texttt{lazyslide\_docs\_retriever} & Search and retrieve information from LazySlide documentation for WSI analysis. \\
\texttt{hovernet\_docs\_retriever} & Search and retrieve information from HoverNet documentation for nuclei segmentation and classification. \\
\bottomrule
\end{tabular}
}
\end{table}

\begin{table}[h]
\centering
\caption{Nuclei segmentation and contour analysis tools available in \ours.}
\label{tab:nuclei-contour-tools}
\vspace{0.5em}
\resizebox{\linewidth}{!}{%
\begin{tabular}{p{0.7\linewidth} p{0.48\linewidth}}
\toprule
\textbf{Tool Name} & \textbf{Description} \\
\midrule
\texttt{segment\_and\_classify\_nuclei\_in\_histology\_roi\_tool} & Segment and classify nuclei in histology ROIs into six classes using HoVer-Net. \\
\texttt{get\_contour\_area} & Calculate the area of a contour. \\
\texttt{get\_contour\_perimeter} & Calculate the perimeter of a contour. \\
\texttt{get\_contour\_convex\_hull} & Calculate the convex hull of a contour. \\
\bottomrule
\end{tabular}
}
\end{table}

\begin{table}[h]
\centering
\caption{Single WSI-level processing and analysis tools available in \ours.}
\label{tab:wsi-analysis-tools}
\vspace{0.5em}
\resizebox{\linewidth}{!}{%
\begin{tabular}{p{0.7\linewidth} p{0.48\linewidth}}
\toprule
\textbf{Tool Name} & \textbf{Description} \\
\midrule
\texttt{visualize\_text\_prompt\_similarity\_on\_wsi\_tool} & Visualize text prompt similarity scores overlaid on WSI tissue regions. \\
\texttt{predict\_wsi\_label\_tool} & Predict WSI-level class labels using text-based zero-shot classification. \\
\texttt{generate\_wsi\_report\_with\_prism\_tool} & Generate a pathology report for a WSI using the PRISM model. \\
\texttt{caption\_single\_wsi\_tool} & Generate descriptive captions for a single WSI by clustering and summarizing. \\
\texttt{score\_tiles\_by\_text\_in\_a\_wsi\_tool} & Score individual tiles in a WSI based on text-based similarity criteria. \\
\texttt{retrieve\_properties\_from\_wsi\_tool} & Retrieve metadata and properties from a single WSI file. \\
\texttt{extract\_tissue\_in\_wsi\_tool} & Perform tissue segmentation on a single WSI file. \\
\texttt{extract\_tissue\_tiles\_in\_wsi\_tool} & Extract tissue tiles/patches from a single WSI file. \\
\texttt{extract\_patch\_features\_in\_wsi\_tool} & Extract patch-level features from a single WSI using foundation models. \\
\texttt{encode\_wsi\_tool} & Encode a single WSI with slide-level features using LazySlide backend. \\
\texttt{check\_tissue\_segmentation\_key\_in\_wsi\_tool} & Check if tissue segmentation results exist for a specific key in WSI. \\
\texttt{check\_tile\_key\_in\_wsi\_tool} & Check if tile extraction results exist for a specific key in WSI. \\
\texttt{check\_patch\_features\_key\_in\_wsi\_tool} & Check if patch features exist for a specific key in WSI. \\
\texttt{check\_slide\_features\_key\_in\_wsi\_tool} & Check if slide-level features exist for a specific key in WSI. \\
\texttt{check\_clustering\_key\_in\_wsi\_tool} & Check if clustering results exist for a specific key in WSI. \\
\texttt{check\_reduction\_key\_in\_wsi\_tool} & Check if dimensionality reduction results exist for a specific key in WSI. \\
\texttt{access\_zarr\_hierarchy} & Access and explore the hierarchical structure of WSI Zarr files. \\
\texttt{read\_zarr\_data\_tool} & Read data from a Zarr file. \\
\texttt{visualize\_wsi\_tool} & Create visualizations of WSI with optional tissue contours and tile overlays. \\
\texttt{reduce\_single\_wsi\_patch\_feature\_space\_tool} & Reduce dimensionality of patch features using PCA, UMAP, or t-SNE. \\
\texttt{run\_leiden\_clustering\_tool} & Perform Leiden clustering on patch features for morphological analysis. \\
\texttt{visualize\_morphological\_clusters\_on\_wsi\_tool} & Visualize morphological clusters overlaid on WSI tissue regions. \\
\texttt{get\_topk\_close\_patch\_coords\_to\_embedding\_space\_clusters\_tool} & Get coordinates of top-k patches closest to embedding space cluster centers. \\
\texttt{read\_rectangle\_region\_from\_wsi\_tool} & Extract rectangular regions from WSI at specified coordinates and magnification. \\
\bottomrule
\end{tabular}
}
\end{table}

\begin{table}[h]
\centering
\caption{WSI classification tools available in \ours.}
\label{tab:wsi-classification-tools}
\vspace{0.5em}
\resizebox{\linewidth}{!}{%
\begin{tabular}{p{0.7\linewidth} p{0.48\linewidth}}
\toprule
\textbf{Tool Name} & \textbf{Description} \\
\midrule
\texttt{train\_test\_wsi\_classification\_mil\_model} & Train and test multiple instance learning models for WSI classification. \\
\texttt{create\_wsi\_classification\_splits} & Create train/validation/test splits for WSI classification datasets. \\
\texttt{prepare\_wsi\_classification\_metadata} & Prepare metadata files for WSI classification experiments. \\
\bottomrule
\end{tabular}
}
\end{table}

\clearpage
\begin{lstlisting}[language=Python,
                   basicstyle=\ttfamily\footnotesize,
                   frame=single,
                   breaklines=true,
                   caption={Example Python tool function: \texttt{dataset\_of\_wsi\_tissue\_segmentation\_tool}},
                   label={fig:wsi-seg}]
def dataset_of_wsi_tissue_segmentation_tool(
    job_dir: str,
    wsi_source: str,
    skip_errors: bool = False,
    search_nested: bool = False,
    holes_are_tissue: bool = True,
    batch_size: int = 64,
    segmentation_model_name: str = 'grandqc',
    tissue_seg_confidence_thresh: float = 0.5,
    overwrite: bool = False,
    skip_specific_wsi: list[str] | None = None,
    keep_only_these_wsi: list[str] | None = None,
    max_workers: int = 16,
) -> dict:
    """
    Run tissue segmentation on multiple WSIs and return the locations of output files.
    Optimized to process multiple WSIs, but can be used with selected WSIs as well.

    Tissue segmentation is the first step for WSI pipelines (patching, feature extraction, etc.).
    Options control whether holes are treated as tissue, thresholding for tissue prediction,
    and artifact removal.

    Notes:
      - If overwrite=True, run segmentation on all valid slides in `wsi_source`.
      - Creates:
          {job_dir}/contours_geojson/{wsi_name}.geojson
          {job_dir}/contours/{wsi_name}.jpg
          {job_dir}/thumbnails/{wsi_name}.jpg
          {job_dir}/_config_segmentation.json
          {job_dir}/_logs_segmentation.txt
      - If overwrite=False, check for existing results and skip processing if found.
      - 'grandqc' performs artifact filtering; 'hest' does not.
      - GeoJSON outputs are GeoPandas GeoDataFrames with `tissue_id` and `geometry`.

    Prerequisites:
      - `job_dir` exists and is writable.
      - `wsi_source` contains valid WSI files.

    Returns (dict):
      - 'dir_with_geojson_contours'
      - 'dir_with_tissue_contours_jpg'
      - 'dir_with_slide_thumbnails'
      - 'tissue_segmentation_log_file'
      - 'tissue_segmentation_config_file'
      - 'number_of_processed_segmentations'
      - 'operation_log'

    Args:
      job_dir: Path to job directory.
      wsi_source: Path to input WSI directory.
      skip_errors: Skip WSIs with errors (default=False).
      search_nested: Recursively search `wsi_source` (default=False).
      holes_are_tissue: Treat holes as tissue (default=True).
      batch_size: Batch size for tile processing (default=64).
      segmentation_model_name: ['grandqc','hest'] (default='grandqc').
      tissue_seg_confidence_thresh: Confidence threshold (default=0.5).
      overwrite: Rerun segmentation if True (default=False).
      skip_specific_wsi: List of WSIs to skip (default=None).
      keep_only_these_wsi: List of WSIs to keep (default=None).
      max_workers: Number of workers (default=16).
    """
\end{lstlisting}

\clearpage
\section{SlideQuest question format}
\label{apd:user-query-format}

Each question in \ourbenchmark follows a unified schema that specifies the task, inputs, and evaluation criteria. The schema enforces consistency across queries while remaining flexible to different modalities and levels of analysis. Core fields include metadata (\texttt{id}, \texttt{data\_type}, \texttt{dataset\_relative\_path}), the main \texttt{question}, and any \texttt{additional\_instructions} or \texttt{output\_instructions} required for reproducibility. Evaluation is standardized through the \texttt{id\_column} and \texttt{columns\_to\_compare\_and\_tolerance}, which define how agent outputs are matched to reference answers. Tolerances may be expressed as numeric thresholds (e.g., allowable percentage error for quantitative tasks) or as sets of acceptable responses for text-based answers. Each query also records a biomedical \texttt{rationale} and verification flags (\texttt{is\_pathologist\_verified}, \texttt{is\_biomedical\_scientist\_verified}) to ensure clinical and scientific validity.
\Cref{tab:user-schema} summarizes the schema fields, and \cref{fig:user-schema-example-dataqa,fig:user-schema-example-cellularqa,fig:user-schema-example-patchqa,fig:user-schema-example-slideqa} shows example instances from each benchmark category. Whenever relevant, question schema's \texttt{additional\_instructions} key specifies to set the seed to 42 for non-deterministic packages (numpy, random, torch, \textit{etc.}) to ensure reproduciliblity between ground truth answers as well as between different runs of the \ours and baselines.

\begin{table}[h]
\centering
\footnotesize
\caption{User query schema used in \ours.}
\label{tab:user-schema}
\vspace{0.5em}
\begin{tabular}{@{} l p{0.65\linewidth} @{}}
\toprule
\textbf{Field} & \textbf{Description} \\
\midrule
\texttt{id} & Unique identifier for the query. \\
\texttt{data\_type} & Type of data involved in the task (e.g., single WSI, multiple WSIs, summary of multiple WSIs). \\
\texttt{dataset\_relative\_path} & Relative path to the dataset. \\
\texttt{question} & The actual biomedical or computational question to be answered. \\
\texttt{additional\_instructions} & Supplementary instructions to provide clarifications. \\
\texttt{output\_instructions} & Explicit requirements on how outputs must be formatted and stored. \\
\texttt{id\_column} & Column name used to compare outputs of agent with ground truth answers. \\
\texttt{columns\_to\_compare\_and\_tolerance} & Fields that are compared and their respective tolerances. \\
\texttt{rationale} & Biomedical motivation or reasoning for why the task is important. \\
\texttt{is\_pathologist\_verified} & Boolean indicating whether the question has been verified by a pathologist. \\
\texttt{is\_biomedical\_scientist\_verified} & Boolean indicating whether the question has been verified by a biomedical scientist. \\
\bottomrule
\end{tabular}
\end{table}

\begin{lstlisting}[language=json,
                   basicstyle=\ttfamily\footnotesize,
                   frame=single,
                   breaklines=true,
                   caption={Example DataQA question schema},
                   label={fig:user-schema-example-dataqa}]
{
    "id": "21",
    "data_type": "single_wsi",
    "slide_relative_path": "tcga_brca_to_use/WSI_flat/TCGA-EW-A1P8-01Z-00-DX1.E9852193-8CDD-49EF-B49B-DA6931198F0D.svs",
    "question": "For the histology slide at {path_to_slide}, what percentage of the tissue pixels are more strongly stained with hematoxylin than with eosin?",
    "additional_instructions": "Your working directory is: {working_dir}, which you can use to save intermediate outputs and results. Work with the lowest resolution level of the whole slide image. You can set non-tissue regions to white before computing stains. Apply 0-1 scaling to each stain matrix before computing percentages. Consider a pixel hematoxylin-dominant if its hematoxylin value exceeds the eosin value by more than 0.02. Report the hematoxylin-dominant tissue percentage to two decimal places.",
    "output_instructions": "You **must** save your outputs as a JSON file in your working directory. Create a file named `answer.json` containing your results as a **list of dictionaries** (JSON array). Use 4-space indentation for readability. The slide name should not include the file extension. hematoxylin_percent **must** be a float rounded to two decimal places. For example, save the following format: [{\"slide_id\": \"slide_id1\", \"hematoxylin_percent\": 44.23}] to the answer.json file with proper indentation.",
    "id_column": "slide_id",
    "columns_to_compare_and_tolerance": {
        "hematoxylin_percent": 0.1
    },
    "rationale": "This calculation quantifies the proportion of tissue dominated by nuclear staining (hematoxylin) versus cytoplasmic/protein staining (eosin), which can indicate cellularity and tissue composition.",
    "is_pathologist_verified": true,
    "is_biomedical_scientist_verified": true
}
\end{lstlisting}

\begin{lstlisting}[language=json,
                   basicstyle=\ttfamily\footnotesize,
                   frame=single,
                   breaklines=true,
                   caption={Example CellularQA question schema},
                   label={fig:user-schema-example-cellularqa}]
{
    "id": "1",
    "data_type": "multiple_wsi",
    "dataset_relative_path": "panoptils_idc_mini/rgbs/",
    "question": "For each breast histology image in {path_to_dataset}, what is the neoplastic cellularity percentage, defined as (number of neoplastic cell nuclei / total number of nuclei) * 100?",
    "additional_instructions": "Your working directory is: {working_dir}, which you can use to save intermediate outputs and results. Don`t consider any background nuclei.",
    "output_instructions": "You **must** save your outputs as a JSON file in your working directory. Create a file named `answer.json` containing your results as a **list of dictionaries** (JSON array). Use 4-space indentation for readability. Percentages should be float values between 0 and 100 with exactly 2 decimal places. For example, if the image name is `sample1.tif` and the neoplastic cellularity is 54.32%, you should save: [{\"slide_id\": \"sample1\", \"neoplastic_cellularity_percentage\": 54.32}] to the answer.json file with proper indentation.",
    "id_column": "slide_id",
    "columns_to_compare_and_tolerance": {
      "neoplastic_cellularity_percentage": 0.15
    },
    "rationale": "Accurate quantification of neoplastic cellularity is essential for assessing neoplastic burden, eligibility for molecular testing, and making informed treatment decisions in oncology.",
    "is_pathologist_verified": true,
    "is_biomedical_scientist_verified": true
}
\end{lstlisting}

\begin{lstlisting}[language=json,
                   basicstyle=\ttfamily\footnotesize,
                   frame=single,
                   breaklines=true,
                   caption={Example PatchQA question schema},
                   label={fig:user-schema-example-patchqa}]
{
    "id": "25",
    "data_type": "multiple_wsi",
    "dataset_relative_path": "panoptils_idc_mini/rgbs/",
    "path_to_metadata": "panoptils_idc_mini/metadata/OS_days/splits.csv",
    "question": "Does adding the average nuclear solidity of neoplastic cells improve Cox proportional hazards model prediction of survival compared to using only clinical covariates (age, stage) in the dataset at {path_to_dataset}? Report the mean C-index IPCW across 5 folds for both conditions",
    "rationale": "Nuclear solidity is a proxy for morphological abnormality; testing its added prognostic value over clinical variables assesses integration of imaging biomarkers with standard clinical data",
    "additional_instructions": "Your working directory is: {working_dir}. Metadata with survival time, event status, clinical covariates (age, stage), patient image paths, and premade splits are in {path_to_metadata} (column `OS` indicates event status, 1 being event observed). To determine the absolute path to the actual image files, join {path_to_dataset} (the dataset root) with the value from the `image_name` column. For each image, compute mean neoplastic nuclear solidity, then average per patient. Using 5-fold CV, train two Cox models: (1) clinical covariates only, (2) clinical covariates and neoplastic nuclear solidity. Compute test-set C-index IPCW per fold, then report mean across folds. Fix random seeds (NumPy, PyTorch, scikit-learn, torch, random, etc.) to 42. A higher risk score indicated by higher values!",
    "output_instructions": "You **must** save your results to a JSON file named `answer.json` in your working directory. The file should contain a list of dictionaries with your outputs. Use json.dump() with indent=4 for readability. Each object must contain `model` (clinical_only/clinical_plus_solidity) and `mean_c_index` (3 decimals). For example: [{\"model\": \"clinical_only\", \"mean_c_index\": 0.652}, {\"model\": \"clinical_plus_solidity\", \"mean_c_index\": 0.687}]",
    "id_column": "model",
    "columns_to_compare_and_tolerance": {
        "mean_c_index": 0.15
    },
    "is_pathologist_verified": true,
    "is_biomedical_scientist_verified": true
}
\end{lstlisting}

\begin{lstlisting}[language=json,
                   basicstyle=\ttfamily\footnotesize,
                   frame=single,
                   breaklines=true,
                   caption={Example SlideQA question schema},
                   label={fig:user-schema-example-slideqa}]
{

    "id": "11",
    "data_type": "multiple_wsi",
    "dataset_relative_path": "tcga_brca_to_use/WSI_flat",
    "path_to_metadata": "tcga_brca_to_use/tcga_brca_metadata/OS_days/metadata.csv",
    "question": "Classify all WSIs at {path_to_dataset} into high and low likelihood of TP53 mutation and report the number of cases in each category. For patients who also have survival data, compare mean overall survival days between the predicted high- and low-likelihood groups and test whether the difference is statistically significant. Consider both censored and uncensored patients.",
    "rationale": "TP53 alterations are linked to prognosis; relating predicted mutation propensity to survival helps validate risk stratification from histology-derived features.",
    "additional_instructions": "Your working directory is: {working_dir}, which you can use to save intermediate outputs and results. Use {path_to_metadata} for patient ids, slide ids, and survival days. Use a two-sided Mann-Whitney U test to compare survival days distributions between groups. Patients missing survival data can be ignored for the survival comparison.",
    "output_instructions": "You must save your results to a JSON file named `answer.json` in your working directory. The file must contain a list of dictionaries (JSON array). Use json.dump() with indent=4 for readability. The dictionary must include numeric keys: `number_high_likelihood` (int), `number_low_likelihood` (int), `avg_survival_days_in_high_likelihood` (float, round to 1 decimal place), `avg_survival_days_in_low_likelihood` (float, round to 1 decimal place), and `p-value` (float, round to 3 decimal places). For example: [{\"number_high_likelihood\": 38, \"number_low_likelihood\": 74, \"avg_survival_days_in_high_likelihood\": 820.5, \"avg_survival_days_in_low_likelihood\": 1160.2, \"p-value\": 0.010}].",
    "id_column": null,
    "columns_to_compare_and_tolerance": {
        "number_high_likelihood": 0.15,
        "number_low_likelihood": 0.15,
        "avg_survival_days_in_high_likelihood": 1.0,
        "avg_survival_days_in_low_likelihood": 1.0,
        "p-value": 0.15
    },
    "is_pathologist_verified": true,
    "is_biomedical_scientist_verified": true
}
\end{lstlisting}

\subsection{\textsc{SlideQuest} evaluation}
\label{apd:slidequest_eval}
To align keys between output and ground truth JSON files, we use Hungarian matching, which minimizes penalties from alternative but valid formatting choices. Values are then compared using task-specific tolerances: percentage thresholds for quantitative outputs (15\% in this study) and sets of acceptable responses for textual outputs. A value is scored as 1 if it falls within tolerance, and 0 otherwise. If no JSON is produced, the question is given a score of 0. Overall benchmark performance is reported as the average score across all questions, including the ones where no output JSON was produced. To define failure rate, we find the percentage of questions where the score is 0 or no valid JSON file is created.

\clearpage
\section{\textsc{SlideQuest} details}
\label{apd:benchmark_details}
\input{tables/capabilities}

\clearpage
\section{Experimental setup}
\label{apd:experiment-setup}

\subsection{\textsc{NOVA} configuration}

\ours is built based on \texttt{smolagents} library. We use the CodeAgent class with the parameters detailed in~\cref{tab:agent_config}. We provide a dynamic set of tools depending on the tools category chosen by the user. \ours with tools, \textit{i.e.,} the default configuration, uses all 49 tools. When evaluating on \ourbenchmark, no baseline and \ours has access to web search. The web search tool is added when evaluating the qualitative case study. Additionally, we increase the maximum number of steps to 200 and memory reset after query to False for the case study to emulate a real interactive conversation with memory.

\begin{table}[h!]
    \centering
    \footnotesize
    \caption{Default \ours configuration parameters}
    \label{tab:agent_config}
    \begin{tabular}{@{} l p{10cm} @{}}
    \toprule
    \textbf{Parameter} & \textbf{Description} \\
    \midrule
    \texttt{tools} & all tools in \ours \\
    \texttt{model} & \texttt{SmolAgentsLLM} configured with \gptfourone by default \\
    \texttt{additional\_authorized\_imports} & list of authorized libraries \\
    \texttt{executor\_type} & local \\
    \texttt{planning\_interval} & null \\
    \texttt{use\_structured\_outputs\_internally} & True \\
    \texttt{verbosity\_level} & 1 \\
    \texttt{provide\_run\_summary} & False \\
    \texttt{max\_steps} & 20 \\
    \texttt{name} &  \texttt{codeagent\_with\_tools}\\
    \midrule
    \texttt{description} & \texttt{The code agent has access to many tools for whole slide image data
      processing and analysis. Additionally, it can use the following libraries (list of libraries).} \\
    \midrule
    \texttt{special\_instructions} & 
\vskip-\baselineskip
\vskip-\smallskipamount
\begin{lstlisting}[basicstyle=\ttfamily\footnotesize,
                   language={},
                   frame=none,
                   breaklines=true,
                   breakindent=1em]
## Security Restrictions
- **Strict restrictions:** You are absolutely not allowed to use these modules in your code: ['os']
## Core Objectives & Approach
- Your primary goal is to help the user fully achieve their objective.
- Always address the user's underlying question or need, not just the surface request. Ensure your answer is complete and fully covers the question and any related aspects.
- **Task Decomposition:** Break down complex tasks into smaller, manageable subtasks and address them sequentially. Execute each subtask one by one, using the appropriate tools and libraries.
## Library & Tool Usage
- **Task Resolution:** Only generate tools and functions from scratch if provided tools and libraries are not able to solve the task.
- **Computer Vision:** When working with image processing, contours, segmentation, or spatial analysis tasks, make use of existing computer vision libraries (cv2, skimage, scipy.spatial and others) before writing custom implementations from scratch.
- The machine you're running in has a gpu. Make sure to always use cuda:0 when a device is required to run a tool.
## Output & Communication
- Display or share outputs---such as figures, files, or results---directly with the user whenever possible
- Exactly follow user instructions on output format, file names, and other details
- Communicate in a sincere, helpful, and user-focused way. Be clear, honest, and avoid unnecessary jargon.
\end{lstlisting}
\kern-\baselineskip
\\
    \bottomrule
    \end{tabular}
\end{table}

\subsection{LLMs variants and parameters}
We use Azure OpenAI endpoints to access the LLMs and benchmark 4 variants of GPT models: \gptfourone, \gptfourmini, \gptfivemini, and \gptfive. The experiments were run within Azure ML, which provides streamlined monitoring and experiment management. For the \gptfourone variants, we used a fixed \texttt{temperature=0} to reduce stochasticity and \texttt{max\_retries=20} to overcome any internal LLM errors. \gptfive series were run with \texttt{temperature=1} as it is currently the only permitted value by the OpenAI API for \gptfive series models. We use the default reasoning configuration for \gptfive. All the baselines were run with the same LLM parameters for fair comparison.

\subsection{Experiments Runner}
We provide an experimental runner to streamline the execution of \ourbenchmark experiments. Built on \texttt{hydra}, it allows dynamic configuration of baselines and system arguments. The runner manages the agent’s lifecycle---including stepwise execution, tool invocation, and result aggregation---while handling evaluation, logging, and saving of intermediate model outputs in separate folders. By resetting the agent’s state and providing a fresh working directory for each query, it prevents leakage between benchmark tasks. This setup also facilitates parallelization, which is particularly useful since each experiment is repeated three times. The runner will be open-sourced as part of the \ours framework.

\clearpage
\input{tables/main_results}

\clearpage
\section{Run time analysis}
\label{apd:run-time-analysis}

\begin{figure}[h]
    \centering
    \includegraphics[width=1.0\linewidth]{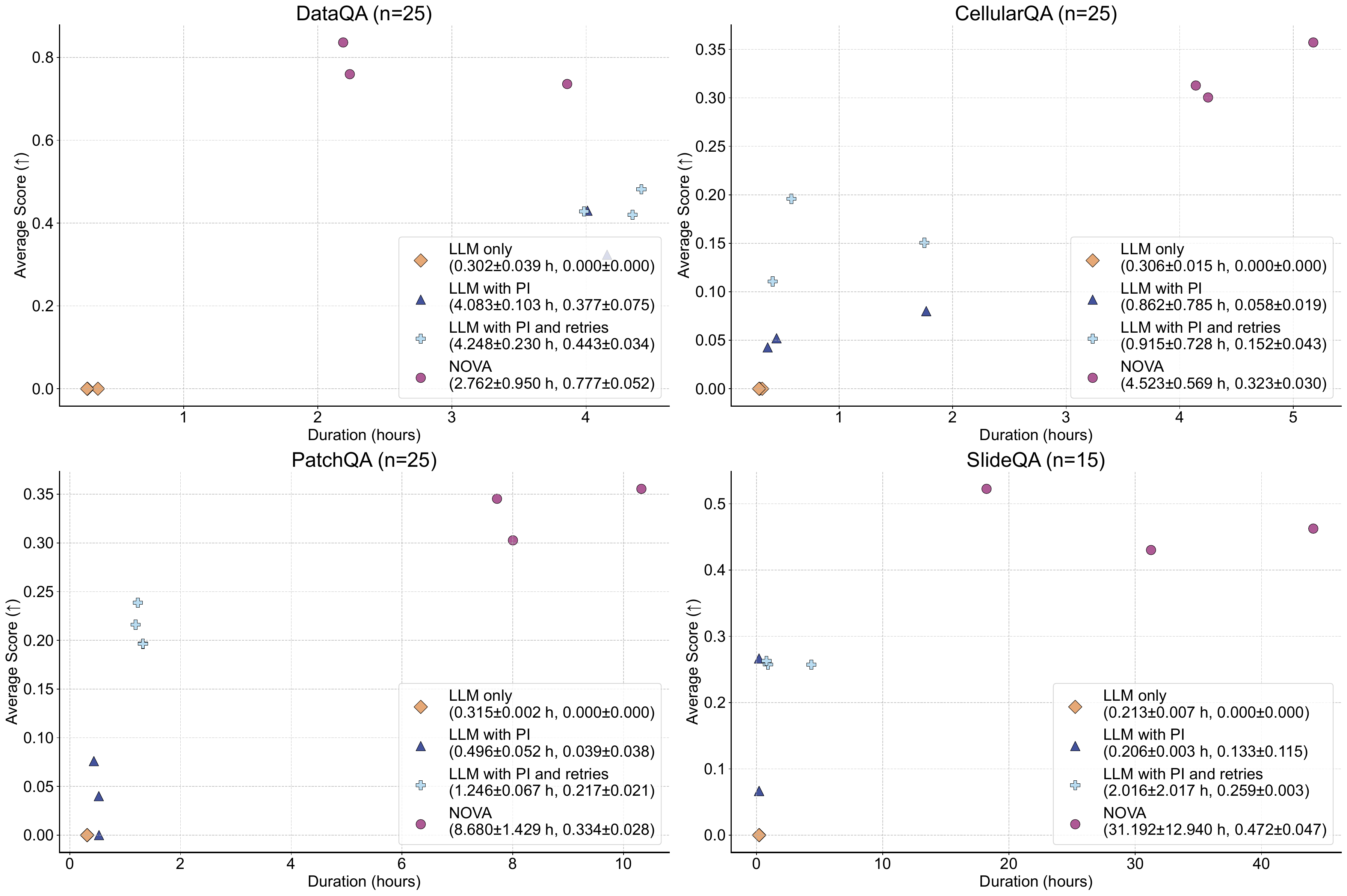}
    \caption{Run time duration (h) vs. average score on \ourbenchmark for \llmonly, \llmwpi, \llmwpir, and \ours. All results with \gptfourone. Legend shows average run time and score with baseline name.}
    \label{fig:duration_baselines}
\end{figure}

\begin{figure}[h]
    \centering
    \includegraphics[width=1.0\linewidth]{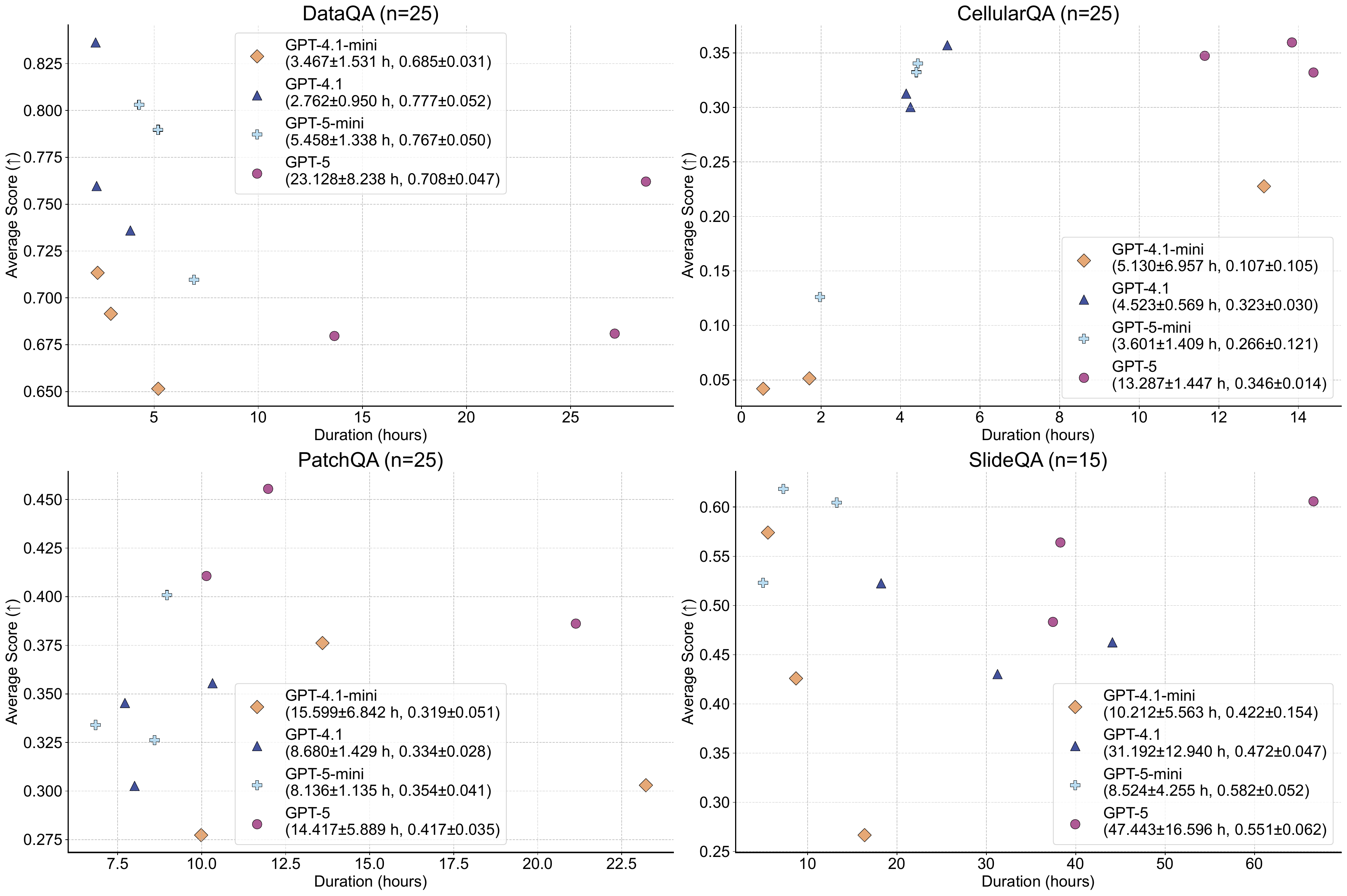}
    \caption{Run time duration (h) vs. average score on \ourbenchmark for \ours with \gptfourmini, \gptfourone, \gptfivemini, and \gptfive. Legend shows average run time and score with baseline name.}
    \label{fig:duration_llms}
\end{figure}

\clearpage
\section{\textsc{NOVA} ablations}
\label{apd:model_ablations}

\subsection{Are custom tools needed?}
\label{sec:rag-baseline}

\ours includes 49 custom tools, each designed and validated by experienced biomedical scientists. To assess whether such handcrafted tools are necessary, we compare \ours against variants where the LLM must generate tools on its own. In the first setting, \ours (no custom tools), the LLM relies solely on its base knowledge to create ad hoc functions.

In the second, \ours (with RAG), the LLM is given access to vector databases built from the GitHub repositories of Trident\footnote{\url{https://github.com/mahmoodlab/TRIDENT}}, LazySlide\footnote{\url{https://github.com/rendeirolab/LazySlide}}, and HoVerNet\footnote{\url{https://github.com/vqdang/hover_net}} (the open-source libraries used for tool development), allowing it to retrieve domain-specific knowledge when creating new tools.
The agent can query these repositories to dynamically generate and compose tools on the fly. This baseline assesses whether access to raw domain-specific resources is sufficient to replace carefully designed tools.

\input{tables/ablations_rag}

\subsection{Choice of core LLM}
\label{sec:llm-quality}
\input{tables/ablations_llm}

\section{\textsc{NOVA} Failure Analysis}
\label{apd:failure_analysis}
We show specific failure cases from \ours highlighting different modes of failure. Question 2 from DataQA (\Cref{fig:failure_case_patch_qa}) shows failure due to tool issues. In solving this task, \ours uses tool \texttt{score\_single\_histology\_image\_using\_text\_tool}, which incorrectly predicts the image class. As an example of the agent ignoring existing tools or data, we look at question 24 from DataQA (\Cref{fig:failure_case_data_qa}), where \ours fails as it incorrectly recomputes the convexity of tissue regions. Specfically, \ours uses the full tissue area when recomputing convexity, instead of using the tissue area without including holes. The convexity measure is already provided by tool \texttt{extract\_tissue\_in\_wsi\_tool} (the docstring also mentioning this), which \ours has already referenced to obtain the tissue segmentation.

\begin{figure}[p]
\centering
(a) Question
\begin{lstlisting}[basicstyle=\ttfamily\footnotesize,
                   language={},
                   frame=single,
                   breaklines=true]
I have three images of a tumour from a breast cancer patient at /.../tcga_ut_brca_zeroshot/TCGA-AR-A2LR.
When considering all the images, does the patient have invasive lobular, invasive ductal, or metaplastic
breast cancer? Additional instructions: ...
\end{lstlisting}

(b) Code (partial)
\begin{lstlisting}[language=Python,
                   basicstyle=\ttfamily\footnotesize,
                   frame=single,
                   breaklines=true]
------------------------------------ Step 2 ------------------------------------
  classes = ['invasive lobular', 'invasive ductal', 'metaplastic']              
  results = []                                           
  for img_path in image_paths: 
      result = score_single_histology_image_using_text_tool(                   
          image_path=img_path,                                                  
          classes=classes,                                                      
          device='cuda:0',                                                      
          apply_softmax=True,                                                   
          prompts=None                                                          
      )                                                                         
      print(f"Image: {img_path}", result)                                       
      results.append(result) 

------------------------------------ Step 3 ------------------------------------
  import re                                                                     
                                                                                
  # Parse similarity scores and sum probabilities for each class                
  class_probs = {cls: 0.0 for cls in classes}                                   
  for result in results:                                                        
      for score_str in result['similarity_scores']:                             
          cls, prob = score_str.split(':')                                      
          cls = cls.strip()                                                     
          prob = float(prob.strip())                                            
          class_probs[cls] += prob                                              
  print(class_probs)                                                            
                                                                                
  # Select the class with the highest summed probability                        
  diagnosis = max(class_probs, key=class_probs.get)                             
  patient_id = Path(patient_folder).name                                        
  output = [{"patient_id": patient_id, "diagnosis": diagnosis}]                 

\end{lstlisting}

(c) Results and ground truth
\begin{lstlisting}[basicstyle=\ttfamily\footnotesize,
                   frame=single,
                   breaklines=true]
Tool output:
    invasive lobular: 0.007
    invasive ductal: 1.981
    metaplastic: 1.012

Agent result: invasive ductal
Ground Truth: metaplastic
\end{lstlisting}

\caption{Question 2 from PatchQA: A failure case as \ours relies on the output of tool \texttt{score\_single\_histology\_image\_using\_text\_tool}, which incorrectly predicts that a majority of images are of type invasive ductal breast cancer and only correctly predicts metaplastic breast cancer for one out of three images.}
\label{fig:failure_case_patch_qa}
\end{figure}

\begin{figure}
\centering
(a) Question
\begin{lstlisting}[basicstyle=\ttfamily\footnotesize,
                   language={},
                   frame=single,
                   breaklines=true]
For histology WSI at /.../tcga_brca_to_use/WSI_flat/TCGA-A2-A0CM-01Z-00-DX1.AC4901DE-4B6D-4185-BB9F
-156033839828.svs, compute the convexity of each tissue region, then return the average convexity per slide.
Work in pixel space! Additional instructions: ...
\end{lstlisting}

(b) Code (partial)
\begin{lstlisting}[language=Python,
                   basicstyle=\ttfamily\footnotesize,
                   frame=single,
                   breaklines=true]
------------------------------------ Step 5 ------------------------------------
# Extract exterior coordinates as a list of [x, y] points                     
polygon = tissues_gdf.geometry.iloc[0]                                        
contour_points = list(polygon.exterior.coords)                                
print(contour_points)                                                         
                                                                            
# Compute area                                                                
area_result = get_contour_area(contour=contour_points)                        
print('Area:', area_result)                                                         
# Compute convex hull                                                         
hull_result = get_contour_convex_hull(contour=contour_points)                 
convex_hull = hull_result['contour_convex_hull']                              
print('Convex hull:', convex_hull)                                            
                                                    
# Compute convex hull area                                                    
hull_area_result = get_contour_area(contour=convex_hull)                      
print('Convex hull area:', hull_area_result)  
\end{lstlisting}

(c) Execution logs
\begin{lstlisting}[basicstyle=\ttfamily\footnotesize,
                   frame=single,
                   breaklines=true]
[...]
Area: {'contour_area': 3084211442.668519}
Convex hull: [...]
Convex hull area: {'contour_area': 3603253238.057195}
\end{lstlisting}

(d) Code (partial)
\begin{lstlisting}[language=Python,
                   basicstyle=\ttfamily\footnotesize,
                   frame=single,
                   breaklines=true]
------------------------------------ Step 6 ------------------------------------
import json                                                                   
slide_id = 'TCGA-A2-A0CM-01Z-00-DX1.AC4901DE-4B6D-4185-BB9F-156033839828'     
area = 3084211442.668519                                                      
hull_area = 3603253238.057195                                                 
convexity = area / hull_area                                                  
avg_convexity = round(convexity, 3)                                           
result = [{"slide_id": slide_id, "avg_convexity": avg_convexity}]             
\end{lstlisting}

(e) Results and ground truth
\begin{lstlisting}[basicstyle=\ttfamily\footnotesize,
                   frame=single,
                   breaklines=true]
Agent result: 0.856
Ground Truth: 1.196
\end{lstlisting}

\caption{Question 24 from DataQA: Failure case as the agent incorrectly recomputes convexity, a value already available through one of the used tools.}
\label{fig:failure_case_data_qa}
\end{figure}

\let\clearpage\relax

\label{LastPage}
\end{document}

%% file: macros.tex
\newcommand{\ours}{\textsc{Nova}\xspace}
\newcommand{\ourbenchmark}{\textsc{SlideQuest}\xspace}

\newcommand{\llmonly}{LLM only\xspace}
\newcommand{\llmwpi}{LLM with PI\xspace}
\newcommand{\llmwpir}{LLM with PI and retries\xspace}

\newcommand{\gptfourone}{\textsc{GPT-4.1}\xspace}
\newcommand{\gptfourmini}{\textsc{GPT-4.1-}mini\xspace}
\newcommand{\gptfive}{\textsc{GPT-5}\xspace}
\newcommand{\gptfivemini}{\textsc{GPT-5-}mini\xspace}

\definecolor{anurag}{rgb}{0.8, 0.2, 0.2}    
\definecolor{steph}{rgb}{0, 0.6, 0.6}   
\definecolor{kenza}{rgb}{0.2, 0.6, 0.2}    
\definecolor{daniel}{rgb}{0.6, 0.4, 0.0}   
\definecolor{felix}{rgb}{0.4, 0.4, 0.8}    
\definecolor{shruthi}{rgb}{0.7, 0.3, 0.7}  
\definecolor{tristan}{rgb}{0.9, 0.6, 0.0}  

\makeatletter
\newcommand{\wip}[1]{{\color{lightgray}#1}\@latex@warning{WIP segment}}
\newcommand{\tbd}[1]{{\color{magenta}#1}\@latex@warning{TBD: #1}}
\newcommand{\@coloredcomment}[2]{{\color{#1}#2}\@latex@warning{#2}}
\newcommand{\TODO}[1]{\@coloredcomment{red}{[TODO: #1]}}
\newcommand{\anuragnote}[1]{\@coloredcomment{anurag}{[Anurag: #1]}}
\newcommand{\stephnote}[1]{\@coloredcomment{steph}{[Steph: #1]}}
\newcommand{\kenzanote}[1]{\@coloredcomment{kenza}{[Kenza: #1]}}
\newcommand{\danielnote}[1]{\@coloredcomment{daniel}{[Daniel: #1]}}
\newcommand{\felixnote}[1]{\@coloredcomment{felix}{[Felix: #1]}}
\newcommand{\shruthinote}[1]{\@coloredcomment{shruthi}{[Shruthi: #1]}}
\newcommand{\tristannote}[1]{\@coloredcomment{tristan}{[Tristan: #1]}}
\makeatother


%% file: tables/capabilities.tex
\begin{table}[h]
\centering
\caption{Numbers of questions requiring each type of capability, per benchmark category.}
\label{tab:capabilities}
\footnotesize
\setlength{\tabcolsep}{3pt}
\sisetup{table-format=2}

\begin{tabular}{@{}llSSSSS@{}}
\toprule
Theme & Capabilities & {CellularQA} & {DataQA} & {PatchQA} & {SlideQA} & {All} \\
\midrule
\multirow[t]{3}{*}{WSI data ingestion}
    & WSI reading & 0 & 12 & 0 & 14 & 26 \\
    & WSI metadata retrieval & 0 & 20 & 0 & 0 & 20 \\
    & WSI patching & 0 & 3 & 0 & 2 & 5 \\
\midrule
\multirow[t]{3}{*}{Basic data ingestion}
    & Image reading & 25 & 0 & 25 & 0 & 50 \\
    & CSV reading & 0 & 0 & 19 & 7 & 26 \\
    & Embeddings reading & 0 & 0 & 0 & 1 & 1 \\
\midrule
\multirow[t]{3}{*}{Image processing}
    & Texture analysis & 2 & 0 & 0 & 0 & 2 \\
    & Image filtering & 0 & 1 & 0 & 0 & 1 \\
    & Stain estimation & 0 & 1 & 0 & 0 & 1 \\
\midrule
\multirow[t]{3}{*}{Spatial analysis}
    & Morphometry & 10 & 4 & 10 & 1 & 25 \\
    & Spatial calibration & 3 & 8 & 2 & 0 & 13 \\
    & Distance computation & 6 & 0 & 1 & 0 & 7 \\
\midrule
\multirow[t]{5}{*}{Statistics}
    & Summary statistics & 23 & 7 & 8 & 4 & 42 \\
    & Statistical analysis & 2 & 0 & 8 & 5 & 15 \\
    & Cross-validation & 0 & 0 & 11 & 3 & 14 \\
    & Model evaluation & 0 & 0 & 10 & 3 & 13 \\
    & Attention-based patch retrieval & 0 & 0 & 0 & 1 & 1 \\
\midrule
\multirow[t]{9}{*}{Pretrained model use}
    & Nuclei cell type classification & 13 & 0 & 12 & 1 & 26 \\
    & Nuclei segmentation & 14 & 0 & 10 & 1 & 25 \\
    & Nuclei detection & 11 & 0 & 2 & 0 & 13 \\
    & Tissue segmentation & 2 & 10 & 0 & 0 & 12 \\
    & Zero-shot WSI interpretation & 0 & 0 & 0 & 10 & 10 \\
    & Patch embedding & 0 & 0 & 7 & 2 & 9 \\
    & Zero-shot multi-patch interpretation & 0 & 0 & 6 & 0 & 6 \\
    & WSI embedding & 0 & 0 & 0 & 3 & 3 \\
    & Artefact segmentation & 0 & 1 & 0 & 0 & 1 \\
\midrule
\multirow[t]{5}{*}{Modelling}
    & Supervised learning & 0 & 0 & 10 & 1 & 11 \\
    & Survival modelling & 0 & 0 & 3 & 0 & 3 \\
    & Few-shot learning & 0 & 0 & 0 & 2 & 2 \\
    & Multiple-instance learning & 0 & 0 & 0 & 2 & 2 \\
    & Clustering & 1 & 0 & 0 & 0 & 1 \\
\midrule
\multirow[t]{2}{*}{General}
    & Scalar calculation & 3 & 6 & 0 & 0 & 9 \\
    & Problem solving & 0 & 0 & 0 & 2 & 2 \\
\bottomrule
\end{tabular}
\end{table}

%% file: tables/main_results.tex
\section{\ourbenchmark Results}
\label{apd:slidequest_results}
\begin{table}[H]
\centering
\caption{\ourbenchmark average score (higher is better) over 3 runs with standard error. Average score is weighted by number of questions in each category. All results with \gptfourone.}
\label{tab:main-results}
\resizebox{\linewidth}{!}{%
\begin{tabular}{>{\raggedright\arraybackslash}m{4cm}
                >{\centering\arraybackslash}m{2.5cm}
                >{\centering\arraybackslash}m{2.5cm}
                >{\centering\arraybackslash}m{2.5cm}
                >{\centering\arraybackslash}m{2.5cm}
                >{\centering\arraybackslash}m{2.5cm}}
\toprule
\textbf{Baseline} & \textbf{DataQA (n=25)} & \textbf{CellularQA (n=25)} & \textbf{PatchQA (n=25)} & \textbf{SlideQA (n=15)} & \textbf{Average} \\
\midrule
\llmonly & $0.000 \pm 0.000$ & $0.000 \pm 0.000$ & $0.000 \pm 0.000$ & $0.000 \pm 0.000$ & $0.000$ \\
\llmwpi & $0.377 \pm 0.053$ & $0.058 \pm 0.011$ & $0.039 \pm 0.022$ & $0.133 \pm 0.067$ & $0.154$ \\
\llmwpir & $0.443 \pm 0.019$ & $0.152 \pm 0.025$ & $0.217 \pm 0.012$ & $0.259 \pm 0.002$ & $0.269$ \\
\rowcolor{gray!20}
\ours & $0.777 \pm 0.030$ & $0.323 \pm 0.017$ & $0.335 \pm 0.016$ & $0.472 \pm 0.027$ & $0.477$ \\
\bottomrule
\end{tabular}%
}

\caption{Failure rate on \ourbenchmark (lower is better) over 3 runs with standard error. Average failure is weighted by number of questions in each category. All results with \gptfourone.}
\label{tab:main-results-failure-perc}
\resizebox{\linewidth}{!}{%
\begin{tabular}{>{\raggedright\arraybackslash}m{4cm}
                >{\centering\arraybackslash}m{2.5cm}
                >{\centering\arraybackslash}m{2.5cm}
                >{\centering\arraybackslash}m{2.5cm}
                >{\centering\arraybackslash}m{2.5cm}
                >{\centering\arraybackslash}m{2.5cm}}
\toprule
\textbf{Baseline} & \textbf{DataQA (n=25)} & \textbf{CellularQA (n=25)} & \textbf{PatchQA (n=25)} & \textbf{SlideQA (n=15)} & \textbf{Average} \\
\midrule
\llmonly & $1.000 \pm 0.000$ & $1.000 \pm 0.000$ & $1.000 \pm 0.000$ & $1.000 \pm 0.000$ & $1.000$ \\
\llmwpi & $0.580 \pm 0.060$ & $0.773 \pm 0.027$ & $0.947 \pm 0.035$ & $0.867 \pm 0.067$ & $0.783$ \\
\llmwpir & $0.507 \pm 0.013$ & $0.627 \pm 0.027$ & $0.613 \pm 0.013$ & $0.667 \pm 0.039$ & $0.596$ \\
\rowcolor{gray!20}
\ours & $0.200 \pm 0.023$ & $0.320 \pm 0.023$ & $0.413 \pm 0.013$ & $0.422 \pm 0.059$ & $0.330$ \\
\bottomrule
\end{tabular}%
}
\end{table}

%% file: tables/ablations_rag.tex
\begin{table}[H]
\centering
\caption{Average score over 3 runs with standard error for \ours (no custom tools), \ours (with RAG), and \ours (with custom tools) on \ourbenchmark (higher is better). Average score is weighted by number of questions in each category. All results with \gptfourone.}
\label{tab:abl-rag}
\vspace{0.5em}
\resizebox{\linewidth}{!}{%
\begin{tabular}{>{\raggedright\arraybackslash}m{4.5cm}
                >{\centering\arraybackslash}m{2.2cm}
                >{\centering\arraybackslash}m{2.2cm}
                >{\centering\arraybackslash}m{2.2cm}
                >{\centering\arraybackslash}m{2.2cm}
                >{\centering\arraybackslash}m{2.2cm}}
\toprule
\textbf{Baseline} & \textbf{DataQA (n=25)} & \textbf{CellularQA (n=25)} & \textbf{PatchQA (n=25)} & \textbf{SlideQA (n=15)} & \textbf{Average} \\
\midrule
\ours (no custom tools) & $0.537 \pm 0.017$ & $0.152 \pm 0.018$ & $0.222 \pm 0.026$ & $0.439 \pm 0.010$ & $0.326$ \\
\ours (with RAG only) & $0.556 \pm 0.006$ & $0.165 \pm 0.018$ & $0.213 \pm 0.015$ & $0.464 \pm 0.013$ & $0.337$ \\
\rowcolor{gray!20}
\ours (with custom tools) & $0.777 \pm 0.030$ & $0.323 \pm 0.017$ & $0.335 \pm 0.016$ & $0.472 \pm 0.027$ & $0.477$ \\
\bottomrule
\end{tabular}%
}

\caption{Failure percentage over 3 runs with standard error for \ours (no custom tools), \ours (with RAG), and \ours (with custom tools) on \ourbenchmark (lower is better). Average failure is weighted by number of questions in each category. All results with \gptfourone.}
\label{tab:abl-rag-failure-perc}
\vspace{0.5em}
\resizebox{\linewidth}{!}{%
\begin{tabular}{>{\raggedright\arraybackslash}m{4.5cm}
                >{\centering\arraybackslash}m{2.2cm}
                >{\centering\arraybackslash}m{2.2cm}
                >{\centering\arraybackslash}m{2.2cm}
                >{\centering\arraybackslash}m{2.2cm}
                >{\centering\arraybackslash}m{2.2cm}}
\toprule
\textbf{Baseline} & \textbf{DataQA (n=25)} & \textbf{CellularQA (n=25)} & \textbf{PatchQA (n=25)} & \textbf{SlideQA (n=15)} & \textbf{Average} \\
\midrule
\ours (no custom tools) & $0.413 \pm 0.013$ & $0.387 \pm 0.035$ & $0.560 \pm 0.061$ & $0.378 \pm 0.022$ & $0.441$ \\
\ours (with RAG only) & $0.400 \pm 0.000$ & $0.573 \pm 0.013$ & $0.573 \pm 0.013$ & $0.400 \pm 0.039$ & $0.496$ \\
\rowcolor{gray!20}
\ours (with custom tools) & $0.200 \pm 0.023$ & $0.320 \pm 0.023$ & $0.413 \pm 0.013$ & $0.422 \pm 0.059$ & $0.330$ \\
\bottomrule
\end{tabular}%
}
\end{table}

%% file: tables/ablations_llm.tex
\begin{table}[H]
\centering
\caption{Performance of different core LLMs on \ourbenchmark. Average score (higher is better) over 3 runs with standard error. Average score is weighted by number of questions in each category. Prices are taken from \url{https://platform.openai.com/docs/pricing} and are the sum of input and output prices per 1M tokens.}
\label{tab:abl-llms}
\vspace{0.5em}
\resizebox{\linewidth}{!}{%
\begin{tabular}{>{\raggedright\arraybackslash}m{3.5cm}
                >{\centering\arraybackslash}m{2.5cm}
                >{\centering\arraybackslash}m{2.2cm}
                >{\centering\arraybackslash}m{2.2cm}
                >{\centering\arraybackslash}m{2.2cm}
                >{\centering\arraybackslash}m{2.2cm}
                >{\centering\arraybackslash}m{2.2cm}}
\toprule
\textbf{Baseline} & \textbf{Price per 1M tokens} & \textbf{DataQA (n=25)} & \textbf{CellularQA (n=25)} & \textbf{PatchQA (n=25)} & \textbf{SlideQA (n=15)} & \textbf{Average} \\
\midrule
\gptfourmini & $\$2.00$ & $0.686 \pm 0.018$ & $0.107 \pm 0.060$ & $0.319 \pm 0.030$ & $0.422 \pm 0.089$ & $0.379$ \\
\rowcolor{gray!20}
\gptfourone & $\$10.00$ & $0.777 \pm 0.030$ & $0.323 \pm 0.017$ & $0.335 \pm 0.016$ & $0.472 \pm 0.027$ & $0.477$ \\
\gptfivemini & $\$2.25$ & $0.767 \pm 0.029$ & $0.266 \pm 0.070$ & $0.354 \pm 0.024$ & $0.582 \pm 0.030$ & $0.482$ \\
\gptfive & $\$11.15$ & $0.708 \pm 0.047$ & $0.346 \pm 0.008$ & $0.417 \pm 0.020$ & $0.551 \pm 0.036$ & $0.498$ \\
\bottomrule
\end{tabular}%
}

\caption{Failure percentage of different core LLMs on \ourbenchmark. Failure percentage (lower is better) over 3 runs with standard error. Average failure percentage is weighted by number of questions in each category. Prices are taken from \url{https://platform.openai.com/docs/pricing} and are the sum of input and output prices per 1M tokens.}
\label{tab:abl-llms-failure-perc}
\vspace{0.5em}
\resizebox{\linewidth}{!}{%
\begin{tabular}{>{\raggedright\arraybackslash}m{3.5cm}
                >{\centering\arraybackslash}m{2.5cm}
                >{\centering\arraybackslash}m{2.2cm}
                >{\centering\arraybackslash}m{2.2cm}
                >{\centering\arraybackslash}m{2.2cm}
                >{\centering\arraybackslash}m{2.2cm}
                >{\centering\arraybackslash}m{2.2cm}}
\toprule
\textbf{Baseline} & \textbf{Price per 1M tokens} & \textbf{DataQA (n=25)} & \textbf{CellularQA (n=25)} & \textbf{PatchQA (n=25)} & \textbf{SlideQA (n=15)} & \textbf{Average} \\
\midrule
\gptfourmini & $\$2.00$ & $0.280 \pm 0.023$ & $0.653 \pm 0.109$ & $0.547 \pm 0.048$ & $0.511 \pm 0.135$ & $0.496$ \\
\rowcolor{gray!20}
\gptfourone & $\$10.00$ & $0.200 \pm 0.023$ & $0.320 \pm 0.023$ & $0.413 \pm 0.013$ & $0.422 \pm 0.059$ & $0.330$ \\
\gptfivemini & $\$2.25$ & $0.213 \pm 0.035$ & $0.373 \pm 0.096$ & $0.373 \pm 0.035$ & $0.289 \pm 0.022$ & $0.315$ \\
\gptfive & $\$11.15$ & $0.270 \pm 0.025$ & $0.280 \pm 0.023$ & $0.373 \pm 0.027$ & $0.244 \pm 0.022$ & $0.297$ \\
\bottomrule
\end{tabular}%
}
\end{table}

%% file: main.bbl
\begin{thebibliography}{41}
\providecommand{\natexlab}[1]{#1}
\providecommand{\url}[1]{\texttt{#1}}
\expandafter\ifx\csname urlstyle\endcsname\relax
  \providecommand{\doi}[1]{doi: #1}\else
  \providecommand{\doi}{doi: \begingroup \urlstyle{rm}\Url}\fi

\bibitem[Adjadj et~al.(2025)Adjadj, Bannier, Horent, Mandela, Lyon, Schutte, Marteau, Gaury, Dumont, Mathieu, et~al.]{adjadj2025towards}
Benjamin Adjadj, Pierre-Antoine Bannier, Guillaume Horent, Sebastien Mandela, Aurore Lyon, Kathryn Schutte, Ulysse Marteau, Valentin Gaury, Laura Dumont, Thomas Mathieu, et~al.
\newblock Towards comprehensive cellular characterisation of h\&e slides.
\newblock \emph{arXiv preprint arXiv:2508.09926}, 2025.

\bibitem[Arora et~al.(2025)Arora, Wei, Hicks, Bowman, Qui{\~n}onero-Candela, Tsimpourlas, Sharman, Shah, Vallone, Beutel, et~al.]{arora2025healthbench}
Rahul~K Arora, Jason Wei, Rebecca~Soskin Hicks, Preston Bowman, Joaquin Qui{\~n}onero-Candela, Foivos Tsimpourlas, Michael Sharman, Meghan Shah, Andrea Vallone, Alex Beutel, et~al.
\newblock Healthbench: Evaluating large language models towards improved human health.
\newblock \emph{arXiv preprint arXiv:2505.08775}, 2025.

\bibitem[Bedi et~al.(2025)Bedi, Cui, Fuentes, Unell, Wornow, Banda, Kotecha, Keyes, Mai, Oez, et~al.]{bedi2025medhelm}
Suhana Bedi, Hejie Cui, Miguel Fuentes, Alyssa Unell, Michael Wornow, Juan~M Banda, Nikesh Kotecha, Timothy Keyes, Yifan Mai, Mert Oez, et~al.
\newblock Medhelm: Holistic evaluation of large language models for medical tasks.
\newblock \emph{arXiv preprint arXiv:2505.23802}, 2025.

\bibitem[Chen et~al.(2025{\natexlab{a}})Chen, Weishaupt, Williamson, Chen, Ding, Chen, Vaidya, Le, Jaume, Lu, et~al.]{chen2025evidence}
Chengkuan Chen, Luca~L Weishaupt, Drew~FK Williamson, Richard~J Chen, Tong Ding, Bowen Chen, Anurag Vaidya, Long~Phi Le, Guillaume Jaume, Ming~Y Lu, et~al.
\newblock Evidence-based diagnostic reasoning with multi-agent copilot for human pathology.
\newblock \emph{arXiv preprint arXiv:2506.20964}, 2025{\natexlab{a}}.

\bibitem[Chen et~al.(2025{\natexlab{b}})Chen, Wang, Ji, Li, Ye, Li, Hu, Yu, Qiao, and He]{chen2025slidechat}
Ying Chen, Guoan Wang, Yuanfeng Ji, Yanjun Li, Jin Ye, Tianbin Li, Ming Hu, Rongshan Yu, Yu~Qiao, and Junjun He.
\newblock Slidechat: A large vision-language assistant for whole-slide pathology image understanding.
\newblock In \emph{Proceedings of the Computer Vision and Pattern Recognition Conference}, pages 5134--5143, 2025{\natexlab{b}}.

\bibitem[Ding et~al.(2025)Ding, Wagner, Song, Chen, Lu, Zhang, Vaidya, Jaume, Shaban, Kim, et~al.]{ding2025multimodal}
Tong Ding, Sophia~J Wagner, Andrew~H Song, Richard~J Chen, Ming~Y Lu, Andrew Zhang, Anurag~J Vaidya, Guillaume Jaume, Muhammad Shaban, Ahrong Kim, et~al.
\newblock A multimodal whole-slide foundation model for pathology.
\newblock \emph{Nature Medicine}, pages 1--13, 2025.

\bibitem[Fallahpour et~al.(2025)Fallahpour, Ma, Munim, Lyu, and Wang]{fallahpour2025medrax}
Adibvafa Fallahpour, Jun Ma, Alif Munim, Hongwei Lyu, and Bo~Wang.
\newblock Medrax: Medical reasoning agent for chest x-ray.
\newblock \emph{arXiv preprint arXiv:2502.02673}, 2025.

\bibitem[Ferber et~al.(2025)Ferber, El~Nahhas, W{\"o}lflein, Wiest, Clusmann, Le{\ss}mann, Foersch, Lammert, Tschochohei, J{\"a}ger, et~al.]{ferber2025development}
Dyke Ferber, Omar~SM El~Nahhas, Georg W{\"o}lflein, Isabella~C Wiest, Jan Clusmann, Marie-Elisabeth Le{\ss}mann, Sebastian Foersch, Jacqueline Lammert, Maximilian Tschochohei, Dirk J{\"a}ger, et~al.
\newblock Development and validation of an autonomous artificial intelligence agent for clinical decision-making in oncology.
\newblock \emph{Nature cancer}, pages 1--13, 2025.

\bibitem[Ghezloo et~al.(2025)Ghezloo, Seyfioglu, Soraki, Ikezogwo, Li, Vivekanandan, Elmore, Krishna, and Shapiro]{ghezloo2025pathfinder}
Fatemeh Ghezloo, Mehmet~Saygin Seyfioglu, Rustin Soraki, Wisdom~O Ikezogwo, Beibin Li, Tejoram Vivekanandan, Joann~G Elmore, Ranjay Krishna, and Linda Shapiro.
\newblock Pathfinder: A multi-modal multi-agent system for medical diagnostic decision-making applied to histopathology.
\newblock \emph{arXiv preprint arXiv:2502.08916}, 2025.

\bibitem[Graham et~al.(2019)Graham, Vu, Raza, Azam, Tsang, Kwak, and Rajpoot]{graham2019hover}
Simon Graham, Quoc~Dang Vu, Shan E~Ahmed Raza, Ayesha Azam, Yee~Wah Tsang, Jin~Tae Kwak, and Nasir Rajpoot.
\newblock Hover-net: Simultaneous segmentation and classification of nuclei in multi-tissue histology images.
\newblock \emph{Medical Image Analysis}, page 101563, 2019.

\bibitem[He et~al.(2020)He, Zhang, Mou, Xing, and Xie]{he2020pathvqa}
Xuehai He, Yichen Zhang, Luntian Mou, Eric Xing, and Pengtao Xie.
\newblock Pathvqa: 30000+ questions for medical visual question answering.
\newblock \emph{arXiv preprint arXiv:2003.10286}, 2020.

\bibitem[He et~al.(2025)He, Li, Liu, Yao, and He]{he2025medorch}
Yexiao He, Ang Li, Boyi Liu, Zhewei Yao, and Yuxiong He.
\newblock Medorch: Medical diagnosis with tool-augmented reasoning agents for flexible extensibility.
\newblock \emph{arXiv preprint arXiv:2506.00235}, 2025.

\bibitem[Heath et~al.(2021)Heath, Ferretti, Agrawal, An, Angelakos, Arya, Bajari, Baqar, Barnowski, Burt, et~al.]{heath2021nci}
Allison~P Heath, Vincent Ferretti, Stuti Agrawal, Maksim An, James~C Angelakos, Renuka Arya, Rosita Bajari, Bilal Baqar, Justin~HB Barnowski, Jeffrey Burt, et~al.
\newblock The nci genomic data commons.
\newblock \emph{Nature genetics}, 53\penalty0 (3):\penalty0 257--262, 2021.

\bibitem[Heng et~al.(2017)Heng, Lester, Tse, Factor, Allison, Collins, Chen, Jensen, Johnson, Jeong, et~al.]{heng2017molecular}
Yujing~J Heng, Susan~C Lester, Gary~MK Tse, Rachel~E Factor, Kimberly~H Allison, Laura~C Collins, Yunn-Yi Chen, Kristin~C Jensen, Nicole~B Johnson, Jong~Cheol Jeong, et~al.
\newblock The molecular basis of breast cancer pathological phenotypes.
\newblock \emph{The Journal of pathology}, 241\penalty0 (3):\penalty0 375--391, 2017.

\bibitem[Jaume et~al.(2024{\natexlab{a}})Jaume, Vaidya, Chen, Williamson, Liang, and Mahmood]{jaume2024modeling}
Guillaume Jaume, Anurag Vaidya, Richard~J Chen, Drew~FK Williamson, Paul~Pu Liang, and Faisal Mahmood.
\newblock Modeling dense multimodal interactions between biological pathways and histology for survival prediction.
\newblock In \emph{Proceedings of the IEEE/CVF Conference on Computer Vision and Pattern Recognition}, pages 11579--11590, 2024{\natexlab{a}}.

\bibitem[Jaume et~al.(2024{\natexlab{b}})Jaume, Vaidya, Zhang, H.~Song, J.~Chen, Sahai, Mo, Madrigal, Phi~Le, and Mahmood]{jaume2024multistain}
Guillaume Jaume, Anurag Vaidya, Andrew Zhang, Andrew H.~Song, Richard J.~Chen, Sharifa Sahai, Dandan Mo, Emilio Madrigal, Long Phi~Le, and Faisal Mahmood.
\newblock Multistain pretraining for slide representation learning in pathology.
\newblock In \emph{European Conference on Computer Vision}, pages 19--37. Springer, 2024{\natexlab{b}}.

\bibitem[Jiang et~al.(2025)Jiang, Black, Geng, Park, Zou, Ng, and Chen]{jiang2025medagentbench}
Yixing Jiang, Kameron~C. Black, Gloria Geng, Danny Park, James Zou, Andrew~Y. Ng, and Jonathan~H. Chen.
\newblock Medagentbench: A virtual ehr environment to benchmark medical llm agents.
\newblock \emph{NEJM AI}, page AIdbp2500144, 2025.
\newblock \doi{10.1056/AIdbp2500144}.
\newblock URL \url{https://ai.nejm.org/doi/full/10.1056/AIdbp2500144}.

\bibitem[Kim et~al.(2024)Kim, Park, Jeong, Chan, Xu, McDuff, Lee, Ghassemi, Breazeal, and Park]{kim2024mdagents}
Yubin Kim, Chanwoo Park, Hyewon Jeong, Yik~S Chan, Xuhai Xu, Daniel McDuff, Hyeonhoon Lee, Marzyeh Ghassemi, Cynthia Breazeal, and Hae~W Park.
\newblock Mdagents: An adaptive collaboration of llms for medical decision-making.
\newblock \emph{Advances in Neural Information Processing Systems}, 37:\penalty0 79410--79452, 2024.

\bibitem[Komura(2022)]{komura2022universal}
D.~Komura.
\newblock Universal encoding of pan-cancer histology by deep texture representations.
\newblock \emph{Cell Reports}, 38:\penalty0 110424, 2022.
\newblock \doi{10.1016/j.celrep.2022.110424}.

\bibitem[Kumar et~al.(2017)Kumar, Verma, Sharma, Bhargava, Vahadane, and Sethi]{kumar2017dataset}
Neeraj Kumar, Ruchika Verma, Sanuj Sharma, Surabhi Bhargava, Abhishek Vahadane, and Amit Sethi.
\newblock A dataset and a technique for generalized nuclear segmentation for computational pathology.
\newblock \emph{IEEE transactions on medical imaging}, 36\penalty0 (7):\penalty0 1550--1560, 2017.

\bibitem[Kumar et~al.(2019)Kumar, Verma, Anand, Zhou, Onder, Tsougenis, Chen, Heng, Li, Hu, et~al.]{kumar2019multi}
Neeraj Kumar, Ruchika Verma, Deepak Anand, Yanning Zhou, Omer~Fahri Onder, Efstratios Tsougenis, Hao Chen, Pheng-Ann Heng, Jiahui Li, Zhiqiang Hu, et~al.
\newblock A multi-organ nucleus segmentation challenge.
\newblock \emph{IEEE transactions on medical imaging}, 39\penalty0 (5):\penalty0 1380--1391, 2019.

\bibitem[Lau et~al.(2018)Lau, Gayen, Ben~Abacha, and Demner-Fushman]{lau2018dataset}
Jason~J Lau, Soumya Gayen, Asma Ben~Abacha, and Dina Demner-Fushman.
\newblock A dataset of clinically generated visual questions and answers about radiology images.
\newblock \emph{Scientific data}, 5\penalty0 (1):\penalty0 1--10, 2018.

\bibitem[Liu et~al.(2024)Liu, Amgad, More, Rathore, Salgado, and Cooper]{liu2024panoptic}
Shangke Liu, Mohamed Amgad, Deeptej More, Muhammad~A Rathore, Roberto Salgado, and Lee~AD Cooper.
\newblock A panoptic segmentation dataset and deep-learning approach for explainable scoring of tumor-infiltrating lymphocytes.
\newblock \emph{NPJ Breast Cancer}, 10\penalty0 (1):\penalty0 52, 2024.

\bibitem[Lu et~al.(2021)Lu, Chen, Williamson, Zhao, Shady, Lipkova, and Mahmood]{lu2021ai}
Ming~Y. Lu, Tiffany~Y. Chen, Drew F.~K. Williamson, Ming Zhao, Mark Shady, Jana Lipkova, and Faisal Mahmood.
\newblock Ai-based pathology predicts origins for cancers of unknown primary.
\newblock \emph{Nature}, 594\penalty0 (7861):\penalty0 106--110, 2021.

\bibitem[Lyu et~al.(2025)Lyu, Liang, Chen, Ding, Yang, Huang, Zhang, He, and Shen]{lyu2025wsi}
Xinheng Lyu, Yuci Liang, Wenting Chen, Meidan Ding, Jiaqi Yang, Guolin Huang, Daokun Zhang, Xiangjian He, and Linlin Shen.
\newblock Wsi-agents: A collaborative multi-agent system for multi-modal whole slide image analysis.
\newblock \emph{arXiv preprint arXiv:2507.14680}, 2025.

\bibitem[Nori et~al.(2023)Nori, King, McKinney, Carignan, and Horvitz]{nori2023capabilities}
Harsha Nori, Nicholas King, Scott~Mayer McKinney, Dean Carignan, and Eric Horvitz.
\newblock Capabilities of gpt-4 on medical challenge problems.
\newblock \emph{arXiv preprint arXiv:2303.13375}, 2023.

\bibitem[Nori et~al.(2025)Nori, Daswani, Kelly, Lundberg, Ribeiro, Wilson, Liu, Sounderajah, Carlson, Lungren, et~al.]{nori2025sequential}
Harsha Nori, Mayank Daswani, Christopher Kelly, Scott Lundberg, Marco~Tulio Ribeiro, Marc Wilson, Xiaoxuan Liu, Viknesh Sounderajah, Jonathan Carlson, Matthew~P Lungren, et~al.
\newblock Sequential diagnosis with language models.
\newblock \emph{arXiv preprint arXiv:2506.22405}, 2025.

\bibitem[Parker et~al.(2009)Parker, Mullins, Cheang, Leung, Voduc, Vickery, Davies, Fauron, He, Hu, et~al.]{parker2009supervised}
Joel~S Parker, Michael Mullins, Maggie~CU Cheang, Samuel Leung, David Voduc, Tammi Vickery, Sherri Davies, Christiane Fauron, Xiaping He, Zhiyuan Hu, et~al.
\newblock Supervised risk predictor of breast cancer based on intrinsic subtypes.
\newblock \emph{Journal of clinical oncology}, 27\penalty0 (8):\penalty0 1160--1167, 2009.

\bibitem[Roucher et~al.(2025)Roucher, del Moral, Wolf, von Werra, and Kaunismäki]{smolagents}
Aymeric Roucher, Albert~Villanova del Moral, Thomas Wolf, Leandro von Werra, and Erik Kaunismäki.
\newblock `smolagents`: a smol library to build great agentic systems.
\newblock \url{https://github.com/huggingface/smolagents}, 2025.

\bibitem[Shaikovski et~al.(2024)Shaikovski, Casson, Severson, Zimmermann, Wang, Kunz, Retamero, Oakley, Klimstra, Kanan, et~al.]{shaikovski2024prism}
George Shaikovski, Adam Casson, Kristen Severson, Eric Zimmermann, Yi~Kan Wang, Jeremy~D Kunz, Juan~A Retamero, Gerard Oakley, David Klimstra, Christopher Kanan, et~al.
\newblock Prism: A multi-modal generative foundation model for slide-level histopathology.
\newblock \emph{arXiv preprint arXiv:2405.10254}, 2024.

\bibitem[Singhal et~al.(2023)Singhal, Azizi, Tu, Mahdavi, Wei, Chung, Scales, Tanwani, Cole-Lewis, Pfohl, et~al.]{singhal2023large}
Karan Singhal, Shekoofeh Azizi, Tao Tu, S~Sara Mahdavi, Jason Wei, Hyung~Won Chung, Nathan Scales, Ajay Tanwani, Heather Cole-Lewis, Stephen Pfohl, et~al.
\newblock Large language models encode clinical knowledge.
\newblock \emph{Nature}, 620\penalty0 (7972):\penalty0 172--180, 2023.

\bibitem[Sun et~al.(2024)Sun, Wu, Zhu, Zheng, Chen, Zhang, Zhang, Wan, Lan, Zheng, et~al.]{sun2024pathmmu}
Yuxuan Sun, Hao Wu, Chenglu Zhu, Sunyi Zheng, Qizi Chen, Kai Zhang, Yunlong Zhang, Dan Wan, Xiaoxiao Lan, Mengyue Zheng, et~al.
\newblock Pathmmu: A massive multimodal expert-level benchmark for understanding and reasoning in pathology.
\newblock In \emph{European Conference on Computer Vision}, pages 56--73. Springer, 2024.

\bibitem[Sun et~al.(2025)Sun, Si, Zhu, Zhang, Shui, Ding, Lin, and Yang]{sun2025cpathagent}
Yuxuan Sun, Yixuan Si, Chenglu Zhu, Kai Zhang, Zhongyi Shui, Bowen Ding, Tao Lin, and Lin Yang.
\newblock Cpathagent: An agent-based foundation model for interpretable high-resolution pathology image analysis mimicking pathologists' diagnostic logic.
\newblock \emph{arXiv preprint arXiv:2505.20510}, 2025.

\bibitem[Tu et~al.(2025)Tu, Schaekermann, Palepu, Saab, Freyberg, Tanno, Wang, Li, Amin, Cheng, et~al.]{tu2025towards}
Tao Tu, Mike Schaekermann, Anil Palepu, Khaled Saab, Jan Freyberg, Ryutaro Tanno, Amy Wang, Brenna Li, Mohamed Amin, Yong Cheng, et~al.
\newblock Towards conversational diagnostic artificial intelligence.
\newblock \emph{Nature}, pages 1--9, 2025.

\bibitem[Vaidya et~al.(2024)Vaidya, Chen, Williamson, Song, Jaume, Yang, Hartvigsen, Dyer, Lu, Lipkova, et~al.]{vaidya2024demographic}
Anurag Vaidya, Richard~J Chen, Drew~FK Williamson, Andrew~H Song, Guillaume Jaume, Yuzhe Yang, Thomas Hartvigsen, Emma~C Dyer, Ming~Y Lu, Jana Lipkova, et~al.
\newblock Demographic bias in misdiagnosis by computational pathology models.
\newblock \emph{Nature Medicine}, 30\penalty0 (4):\penalty0 1174--1190, 2024.

\bibitem[Vaidya et~al.(2025)Vaidya, Zhang, Jaume, Song, Ding, Wagner, Lu, Doucet, Robertson, Almagro-Perez, et~al.]{vaidya2025molecular}
Anurag Vaidya, Andrew Zhang, Guillaume Jaume, Andrew~H Song, Tong Ding, Sophia~J Wagner, Ming~Y Lu, Paul Doucet, Harry Robertson, Cristina Almagro-Perez, et~al.
\newblock Molecular-driven foundation model for oncologic pathology.
\newblock \emph{arXiv preprint arXiv:2501.16652}, 2025.

\bibitem[Wang et~al.(2024)Wang, Chen, Yuan, Zhang, Li, Peng, and Ji]{wang2024executable}
Xingyao Wang, Yangyi Chen, Lifan Yuan, Yizhe Zhang, Yunzhu Li, Hao Peng, and Heng Ji.
\newblock Executable code actions elicit better llm agents.
\newblock In \emph{Forty-first International Conference on Machine Learning}, 2024.

\bibitem[W{\"o}lflein et~al.(2025)W{\"o}lflein, Ferber, Truhn, Arandjelovi{\'c}, and Kather]{wolflein2025llm}
Georg W{\"o}lflein, Dyke Ferber, Daniel Truhn, Ognjen Arandjelovi{\'c}, and Jakob~Nikolas Kather.
\newblock Llm agents making agent tools.
\newblock \emph{arXiv preprint arXiv:2502.11705}, 2025.

\bibitem[Zhang et~al.(2025)Zhang, Jaume, Vaidya, Ding, and Mahmood]{zhang2025standardizing}
Andrew Zhang, Guillaume Jaume, Anurag Vaidya, Tong Ding, and Faisal Mahmood.
\newblock Accelerating data processing and benchmarking of ai models for pathology.
\newblock \emph{arXiv preprint arXiv:2502.06750}, 2025.

\bibitem[Zheng et~al.(2025)Zheng, Abila, Chrenkov{\'a}, Winkler, and Rendeiro]{zheng2025lazyslide}
Yimin Zheng, Ernesto Abila, Eva Chrenkov{\'a}, Juliane Winkler, and Andr{\'e}~F Rendeiro.
\newblock Lazyslide: accessible and interoperable whole slide image analysis.
\newblock \emph{BioRxiv}, pages 2025--05, 2025.

\bibitem[Zhu et~al.(2025)Zhu, He, Hu, Zheng, Zhang, Wang, Gao, Ma, and Yu]{zhu2025medagentboard}
Yinghao Zhu, Ziyi He, Haoran Hu, Xiaochen Zheng, Xichen Zhang, Zixiang Wang, Junyi Gao, Liantao Ma, and Lequan Yu.
\newblock Medagentboard: Benchmarking multi-agent collaboration with conventional methods for diverse medical tasks.
\newblock \emph{arXiv preprint arXiv:2505.12371}, 2025.

\end{thebibliography}
